    \documentclass[times, review, 10pt]{elsarticle}
    
   \usepackage[margin=1in]{geometry}
    
    \usepackage{amssymb}
    \usepackage{amsmath}
    \usepackage{amsthm}
    \usepackage{booktabs}
    \usepackage{subfigure}
    \usepackage{lineno}
    \usepackage{float}
    \usepackage{tikz}
    \usetikzlibrary{shapes.geometric, arrows.meta, positioning}
    \usepackage{graphicx}
    \usepackage{multirow}
    \usepackage{mathrsfs}
    \usepackage[title]{appendix}
    \usepackage{textcomp}
    \usepackage{manyfoot}
    \usepackage{algorithm}
    \usepackage{subcaption}
    \usepackage{algorithmicx}
    \usepackage{algpseudocode}
    \usepackage{listings}
    \usepackage{comment}
    \usepackage{hyperref}
    \usepackage{cleveref}
    \usepackage{xcolor}

    \usepackage{setspace}
    \usepackage{lineno}
    
    \journal{}
    
\begin{document}
    
    \begin{frontmatter}

    \title{RhyMix: A Lightweight Adaptive Multi-Rhythm Network for Long-Term Time Series Forecasting}
    
    \author[1]{Sumit Satishrao Shevtekar\textsuperscript{*}}
    \ead{phd2401101010@iiti.ac.in}   
    \author[1]{Chandresh Kumar Maurya}
    \ead{chandresh@iiti.ac.in}
    
    \address[1]{Department of Computer Science and Engineering, Indian Institute of Technology Indore, Indore, 453552, Madhya Pradesh, India}

    \cortext[cor1]{Corresponding author}
    
    
    \begin{abstract}
Real-world time series exhibit complex dynamics characterized by multiple simultaneous temporal patterns: short-term fluctuations, periodic seasonal cycles, long-term trends, and irregular abrupt changes. However, most forecasting architectures rely on single-path temporal modeling---transformers capture long-range dependencies but smooth local variations, convolutions capture local patterns but have limited receptive fields, and linear models are efficient but cannot capture nonlinear dynamics---so no single architectural paradigm captures all these pattern types simultaneously. To address this, we introduce \emph{RhyMix (RHYthm MIXture)}, a hybrid neural architecture designed around a parallel dual-path modeling paradigm with adaptive gating mechanisms. RhyMix integrates two complementary encoding branches: (i) a \emph{Cyclic Path} that incorporates explicit seasonal inductive bias through learnable cyclic embeddings; and (ii) a lightweight \emph{Multi-Scale Temporal Convolutional Network with Channel Attention (MSTCN-CA) Path} that captures temporal dependencies across different receptive fields. Unlike prior hybrid architectures that combine components with fixed weights, RhyMix computes its fusion weights from the input, adapting to each sample. Its key methodological contribution is a \emph{two-level adaptive gating framework}: a per-channel path gate that fuses four specialized forecasting heads, and a per-sample hybrid gate that balances the two paths. This design maintains linear complexity. Across 12 real-world benchmarks, RhyMix achieves the best average MSE on 9 datasets and the best average MAE on 8 datasets. The model remains lightweight (40K parameters for $C{=}7$) with low-latency inference (5.2--7.5 ms), making it suitable for resource-constrained edge devices and real-time deployment.
\end{abstract}
    



    

    \end{frontmatter}
    
    


    \section{Introduction}\label{sec1}

    Multivariate time series (MTS) analysis underpins a broad spectrum of societally and industrially critical applications, including energy systems, intelligent transportation systems, weather forecasting, industrial prognostics, and behavioral analytics. Across these domains, learning from temporal data is central to tasks such as long-term forecasting, missing-value imputation, and anomaly detection~\citep{wu2022autoformer, liu2024itransformerinvertedtransformerseffective, WANG2024110118, HOU2025111037}. Despite decades of study, robust temporal modeling remains fundamentally challenging. Unlike language or vision, individual time points in a time series carry limited semantic meaning; useful information emerges only through temporal variation patterns such as continuity, periodicity, and long-term trends~\citep{wu2023timesnettemporal2dvariationmodeling, lim2021time}. This intrinsic dependency structure renders time series modeling both information-rich and computationally demanding.
    
    Deep learning (DL) has substantially advanced time series analysis, not only improving predictive accuracy but also enabling transferable representations for downstream tasks~\citep{nie2023timeseriesworth64}. However, recent evidence that simple linear models can rival or even surpass sophisticated Transformer-based architectures on standard benchmarks~\citep{zeng2022transformerseffectivetimeseries} has exposed structural limitations in contemporary designs. Classical neural architectures— including convolutional networks~\citep{franceschi2019unsupervised} and recurrent models with gating mechanisms~\citep{6795963,lai2018modelinglongshorttermtemporal}—are constrained by vanishing gradients, limited parallelism, or restricted receptive fields. Although temporal convolutional networks improve efficiency~\citep{he2019temporal}, they rely on fixed receptive scales. Transformer-based models promise global dependency modeling, yet their quadratic complexity in sequence length has driven a variety of approximate attention mechanisms.~\citep{vaswani2023attentionneed}. Many such methods continue to treat time steps as independent tokens, limiting their ability to encode local semantic structure and multi-scale temporal interactions.

    These design pressures have led to the emergence of three dominant modeling paradigms, formalized as Channel Strategies~\citep{qiu2025comprehensivesurveydeeplearning}: Channel Independence (CI), Channel Dependence (CD), and Channel Partiality (CP). CI prioritizes efficiency but sacrifices cross-channel interactions. CD captures dependencies but scales poorly. CP seeks compromise via latent projections. All rely on rigid structural assumptions—fixed patch sizes, uniform MLPs, static backbones—limiting adaptability. As a result, even state-of-the-art (SOTA) systems such as PatchTST~\citep{nie2023timeseriesworth64}, TimesNet~\citep{wu2023timesnettemporal2dvariationmodeling}, iTransformer~\citep{liu2024itransformerinvertedtransformerseffective}, TSMixer~\citep{chen2023tsmixer}, SOFTS~\citep{han2024softsefficientmultivariatetime}, recent LLM-inspired approaches~\citep{10.1145/3719207, jin2024timellmtimeseriesforecasting} and even the recently proposed GCMNet~\citep{ LIU2026113287} struggle to maintain temporal fidelity under complex, real-world conditions. 

    Recently, Time-o1~\citep{wang2026timeo} introduced a transformation-augmented learning objective that addresses label autocorrelation and optimization challenges through SVD-based projection. While Time-o1 achieves competitive performance across multiple benchmarks, it adds a projection matrix $P^* \in \mathbb{R}^{H \times H}$ requiring SVD pre-computation, resulting in quadratic complexity $\mathcal{O}(H^2)$ with respect to horizon. Furthermore, Time-o1 requires a base model (e.g., iTransformer with $\sim$400K parameters), significantly increasing the overall parameter count to $\sim$409K at H=96. This dependency on external base models and pre-computation limits its deployability in resource-constrained environments. Other recent approaches face complementary limitations. MSTN~\citep{shevtekar2026mstn}, a hybrid CI+CD model, employs early temporal aggregation that may discard fine-grained information. CycleNet~\citep{lin2024cyclenetenhancingtimeseries} relies on a single period. ModernTCN~\citep{donghao2024moderntcn} lacks explicit seasonal priors. 
    
    A key limitation shared by existing forecasting approaches is the reliance on a single temporal modeling strategy, despite real-world time series exhibiting multiple pattern types simultaneously—short-term fluctuations, long-term trends, and repeating cycles. This limitation is architecturally evident in methods such as TSMixer and DLinear, which combine predictors with fixed weights, ignoring that the optimal combination varies with input characteristics.  These approaches assume a single dominant scale, ignore sample-specific variations, and often require millions of parameters. Consequently, such architectures face challenges in real-world settings where multi-scale dynamics and sample-specific characteristics are present. There is a need for architectures that are not only effective for long-term forecasting but also lightweight and efficient.

      Towards this end, we introduce \emph{RhyMix}, a hybrid neural architecture designed around a parallel dual-path modeling paradigm with adaptive gating mechanisms. Unlike models that rely on a single modeling approach, RhyMix integrates two complementary encoding branches: (i) a \emph{Cyclic Path} that incorporates explicit seasonal inductive bias through learnable cyclic embeddings capturing predictable rhythmic patterns; and (ii) a lightweight \emph{Multi-Scale Temporal Convolutional Network (TCN) with Channel Attention (MSTCN-CA) Path} that employs multi-scale depthwise dilated convolutions to capture temporal dependencies across different receptive fields. A key design element is the use of \emph{adaptive gating mechanisms} at multiple levels: a path gate dynamically combines four specialized forecasting heads (Direct, Trend-Seasonal Decomposition, Local Convolution, and Periodic Fusion) for each sample and channel, while a hybrid gate adaptively balances the contributions of the Cyclic and MSTCN-CA Path based on input characteristics. This design ensures the model adapts its behavior to the specific temporal patterns present in each sample while maintaining computational efficiency through linear complexity $\mathcal{O}(L \cdot C)$ with respect to sequence length and number of channels.

   We evaluate RhyMix on twelve publicly available benchmark datasets for long-term forecasting. These cover diverse domains including electricity, traffic, weather, and finance.    The results demonstrate that RhyMix achieves the best average MSE on 9 datasets and the best average MAE on 8 datasets, while using 7--16$\times$ fewer parameters than Time-o1 and TimeMixer++ on low-dimensional datasets and requiring no pre-computation. On ETTh1 ($C=7$), RhyMix achieves 5.85 ms inference latency, 157 KB model size, and 10.9 MB inference memory, making it suitable for resource-constrained edge deployment.

    Our contributions in this work are summarized as follows:
    
   \begin{enumerate}

    \item We introduce \emph{RhyMix}, a lightweight hybrid neural architecture combining 
    explicit multi-period cyclic modeling with multi-scale TCN-based temporal feature extraction through a parallel dual-path design. Its key methodological contribution is a \emph{two-level adaptive gating framework} — a per-channel path gate fusing four forecasting heads and a per-sample hybrid gate fusing the two paths based on the input, with linear complexity 
    $\mathcal{O}(L \cdot C \cdot (K + D + r + 4H))$.
          
    \item Through extensive evaluation across \textbf{12} benchmark datasets, RhyMix achieves the best average MSE on \textbf{9} datasets and the best average MAE on \textbf{8} datasets, outperforming recent models such as TimeMixer++, Time-o1, GCMNet, SEG-MOE, SOFTS, MSTN, FilterNet (TexFilter), iTransformer, PatchTST, and TSMixer.
    
    \item RhyMix demonstrates practical deployability with \textbf{40,269 parameters} for low-dimensional datasets ($C=7$, 157 KB model size, 5.85 ms latency, 10.9 MB inference memory), making it suitable for resource-constrained edge devices and real-time applications.
    
    \end{enumerate}

\section{Related Work}
\label{sec:RelatedWork}
In multivariate time series forecasting, substantial research has been conducted. However, we review recent advances in long-term time series forecasting, focusing on two key aspects: (i) architectural strategies for handling multivariate channels, and (ii) computational efficiency and resource requirements.
    
\subsection{Architectural Strategies for Multivariate Channels}
In this section, We focus on long-term forecasting, the primary focus of RhyMix.
        
The majority of recent works focus on long-term forecasting tasks, including specialized architectures such as PatchTST~\citep{nie2023timeseriesworth64}, iTransformer~\citep{liu2024itransformerinvertedtransformerseffective}, TSMixer~\citep{chen2023tsmixer}, TIME-LLM~\citep{jin2024timellmtimeseriesforecasting}, and LLM4TS~\citep{10.1145/3719207}. SEG-MOE~\citep{ortigossa2026segmoemultiresolutionsegmentwisemixtureofexperts} improves scalability and long-term dependency modeling via segment-wise Mixture-of-Experts routing. SOFTS~\citep{han2024softsefficientmultivariatetime} introduces an efficient MLP-based architecture with a Star Aggregate-Dispatch module that aggregates all series into a global core representation. FilterNet (TexFilter)~\citep{yi2024filternet} leverages frequency-domain filtering to capture periodic patterns. DLinear~\citep{zeng2022transformerseffectivetimeseries} decomposes time series into trend and seasonal components with simple linear layers. TimesNet~\citep{wu2023timesnettemporal2dvariationmodeling} transforms 1D time series into 2D representations to capture complex temporal patterns. Time-o1~\citep{wang2026timeo} introduces a transformation-augmented learning objective that adds a projection matrix $P^* \in \mathbb{R}^{H \times H}$ requiring SVD pre-computation. Other designs such as DPIA~\citep{TAN2026114310}, CVACL-MA~\citep{YE2026111936}, visibility graph methods~\citep{ZHAN2024110720}, and parallel multi-scale dynamic GNNs~\citep{HOU2025111037} address complementary challenges including pattern-level interaction, variate decomposition, complex network extraction, and multi-scale graph learning. These models have driven significant performance gains in long-term forecasting.

The works reviewed above handle channel dependency differently, following the established CI, CD, or CP taxonomy~\citep{qiu2025comprehensivesurveydeeplearning}. CI methods such as DLinear~\citep{zeng2022transformerseffectivetimeseries}, PatchTST~\citep{nie2023timeseriesworth64}, and CycleNet~\citep{lin2024cyclenetenhancingtimeseries} treat each channel independently and therefore cannot model cross-channel effects. CD models such as iTransformer~\citep{liu2024itransformerinvertedtransformerseffective} capture full cross-channel dependencies but scale poorly with channel count, often with $\mathcal{O}(C^2)$ complexity. CP methods like MCformer~\citep{10533212} and ModernTCN~\citep{donghao2024moderntcn} project the channel dimension into a latent space before interaction, trading expressivity for efficiency. While these methods show impressive performance, they are often limited in capturing multi-scale temporal dynamics across different frequencies—a capability essential for modeling long-sequence dependencies with varying periodicities.
    
Recent hybrid approaches have emerged to balance the strengths of CI and CD paradigms. MSTN~\citep{shevtekar2026mstn} employs a dual-encoder architecture with parallel CNN and sequence modeling branches (Transformer or BiLSTM), applying early temporal aggregation to collapse the sequence length before fusion, achieving $\mathcal{O}(1)$ complexity in refinement stages. MSTN demonstrates SOTA performance across imputation, forecasting, classification, and cross-dataset generalization while remaining lightweight ($\sim$0.40M params for MSTN-BiLSTM and $\sim$1.06M for MSTN-Transformer). However, MSTN relies on Transformer or BiLSTM encoders with $\mathcal{O}(L^2)$ or $\mathcal{O}(L)$ complexity respectively, and its ETA mechanism, while effective for efficiency, may discard fine-grained temporal information before multi-scale fusion. 

To combine CI and CD strengths without heavy encoders, RhyMix uses a dual-path design: the Cyclic Path processes each channel independently via multi-period cyclic tables (12, 24, 48, 168), while the MSTCN-CA Path follows CI through depthwise convolutions with selective CD interaction via a lightweight channel mixer. Unlike MSTN, it maintains $\mathcal{O}(L \cdot C)$ complexity without early aggregation. Unlike CycleNet's single period, it captures multi-scale seasonal patterns through multiple periods, balancing channel independence and cross-channel dependencies without quadratic cost.

\subsection{Computational Efficiency and Resource Requirements}

In this section, we discuss the efficiency of time series models in terms of (i) characteristics of time series captured such as periodicity, varying temporal patterns, and non-stationary behaviors; (ii) inference latency; and (iii) memory footprint.

CNN-based approaches model temporal patterns through convolutional operations. TimesNet~\citep{wu2023timesnettemporal2dvariationmodeling} (2D CNNs) reshapes time series into 2D images to capture complex temporal variations. Transformer-based approaches include Informer, which reduces attention 
complexity via sparse selection, Autoformer~\citep{wu2022autoformer}, and 
FEDformer~\citep{zhou2022fedformer}, which uses autocorrelation-based decomposition to capture periodic patterns, achieving strong results on long-horizon forecasting. Nevertheless, their reliance on stable periodic assumptions makes them vulnerable to irregular events and abrupt changes. Sparse and frequency-domain attention variants reduce some computational overhead, yet they still operate over the full input length, leading to 
growing memory and runtime costs as the sequence grows. They also tend to specialize in either local or global trends, rarely balancing both effectively while meeting the requirement for low-latency inference.

MLP-based designs such as SOFTS~\citep{han2024softsefficientmultivariatetime}, TSMixer~\citep{chen2023tsmixer}, and TTM~\citep{ekambaram2024tinytimemixersttms} offer greater computational efficiency through temporal and channel mixing, often outperforming attention-based models in speed. However, their mixing operations are static and do not adapt to the varying temporal characteristics across different samples or frequencies. Another line of work leverages large language models for forecasting, exemplified by LLM4TS~\citep{10.1145/3719207}, which exploit few-shot learning capabilities but incur prohibitive latency and memory overhead, making them unsuitable for real-time edge deployment.

In terms of scalability, iTransformer~\citep{liu2024itransformerinvertedtransformerseffective} captures full cross-channel dependencies but scales poorly with channel count, often with $\mathcal{O}(C^2)$ complexity, as it treats each variate as a token and applies self-attention across channels. CycleNet~\citep{lin2024cyclenetenhancingtimeseries} introduces cyclic priors using learnable periodic tables but relies on a single period and fails to capture multi-scale patterns. ModernTCN~\citep{donghao2024moderntcn} employs temporal convolutional networks with multi-scale dilations but lacks explicit seasonal priors. MSTN~\citep{shevtekar2026mstn} achieves SOTA performance with early temporal aggregation, enabling $\mathcal{O}(1)$ refinement stages, but its front-end encoder retains $\mathcal{O}(L^2)$ or $\mathcal{O}(L)$ complexity, and early aggregation may discard fine-grained temporal details. Time-o1~\citep{wang2026timeo} requires a base model and adds a projection matrix $P^* \in \mathbb{R}^{H \times H}$ with $\mathcal{O}(H^2)$ complexity. These limitations highlight the need for architectures that deliver improved long-term forecasting performance while remaining lightweight and efficient. RhyMix attempts to address this gap by combining explicit multi-period cyclic priors with multi-scale dilated convolutions under adaptive gating, achieving linear complexity throughout. 
    
\section{Methodology}
    
This section details the RhyMix architecture and its design rationale.

\subsection{Proposed Architecture: RhyMix}
\label{RhyMix}
    
\subsubsection{Architecture Overview}
\label{Arch_overview}

\begin{figure}[t]
\centering
\includegraphics[width=0.99\textwidth]{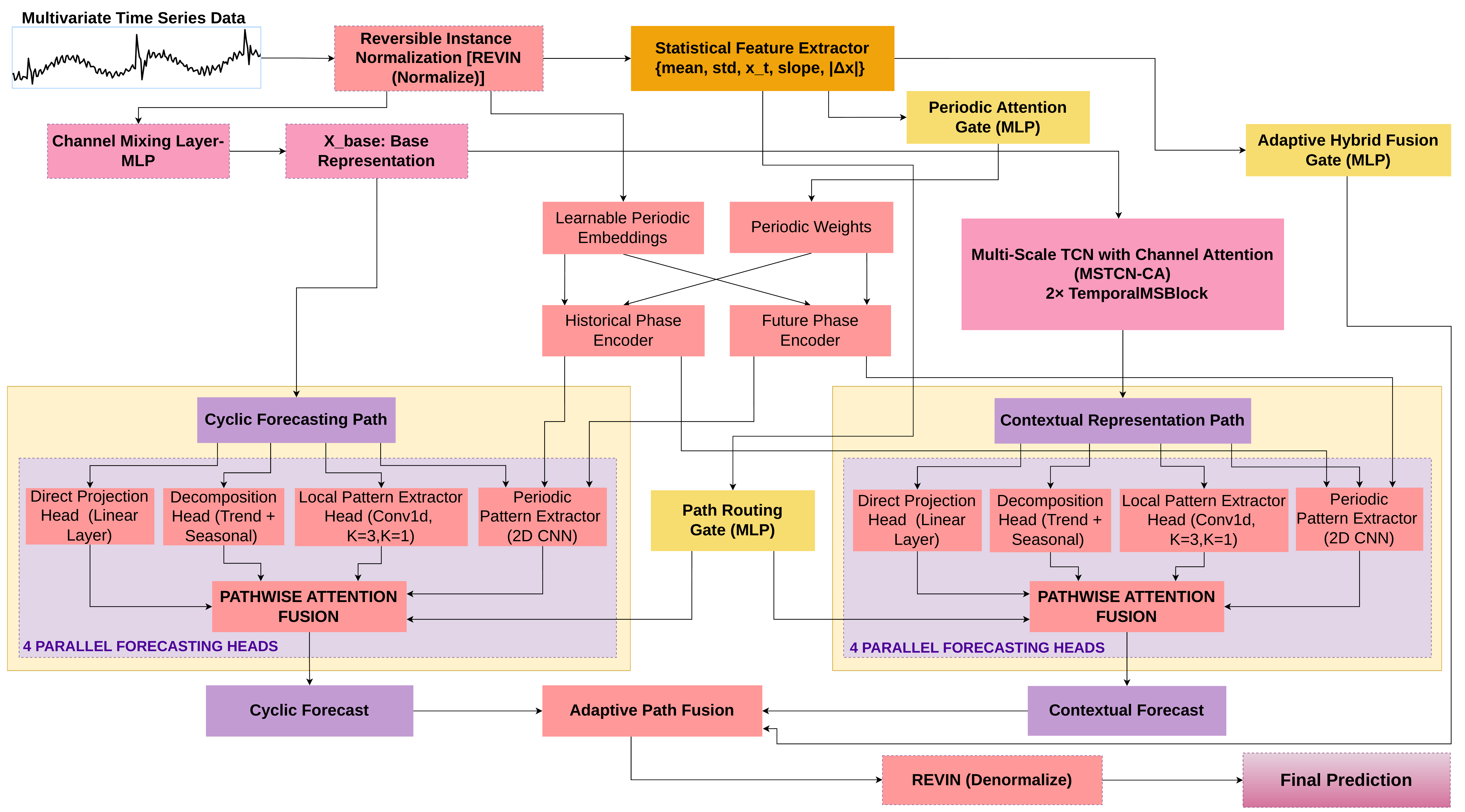}
\caption{RhyMix architecture.}
\label{fig:ArchDiagram}
\end{figure}

The proposed \textit{RhyMix} (Fig.~\ref{fig:ArchDiagram}) is a hybrid deep-learning architecture designed around a parallel dual-path modeling paradigm. RhyMix processes temporal signals through two coordinated encoding branches, each specialized for complementary aspects of temporal structure: explicit seasonal patterns and data-driven multi-scale dynamics. A key design element is the use of adaptive gating mechanisms applied at multiple levels throughout the network, which dynamically modulates the contributions of different components based on input characteristics. This design ensures that the model adapts its behavior to the specific temporal patterns present in each sample, while maintaining computational efficiency through linear complexity with respect to sequence length.

\begin{figure}
\centering
\includegraphics[width=0.95\textwidth]{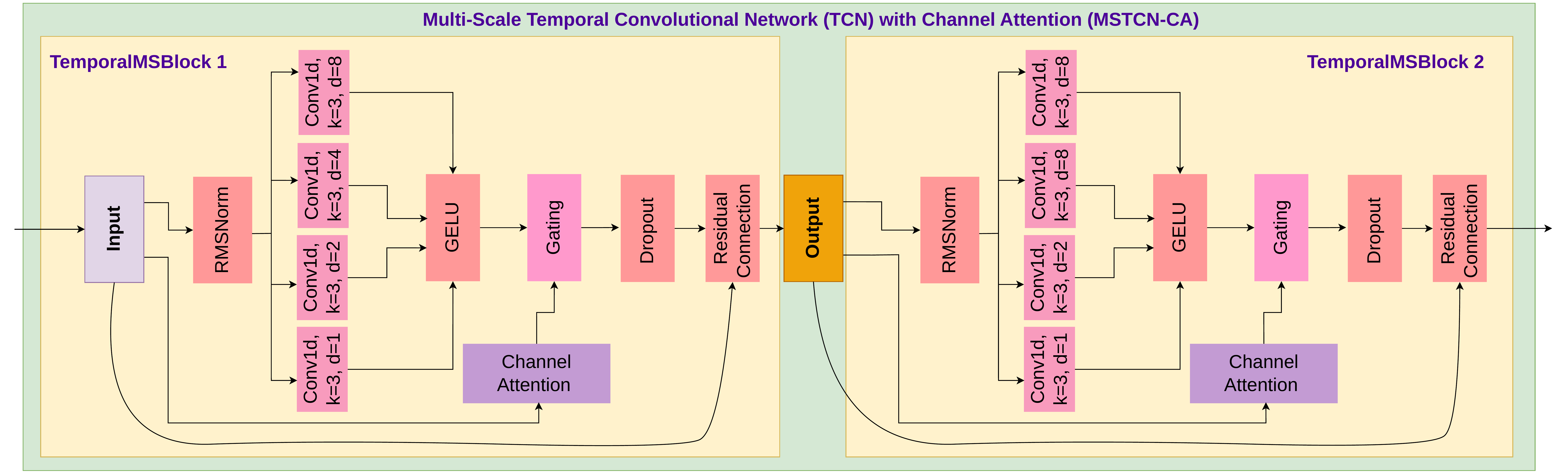}
\caption{Architecture of MSTCN-CA with two TMS blocks, (d=1,2,4,8) and channel attention.}
\label{fig:mstcn_ca}
\end{figure}

\begin{figure}
\centering
\includegraphics[width=0.85\textwidth]{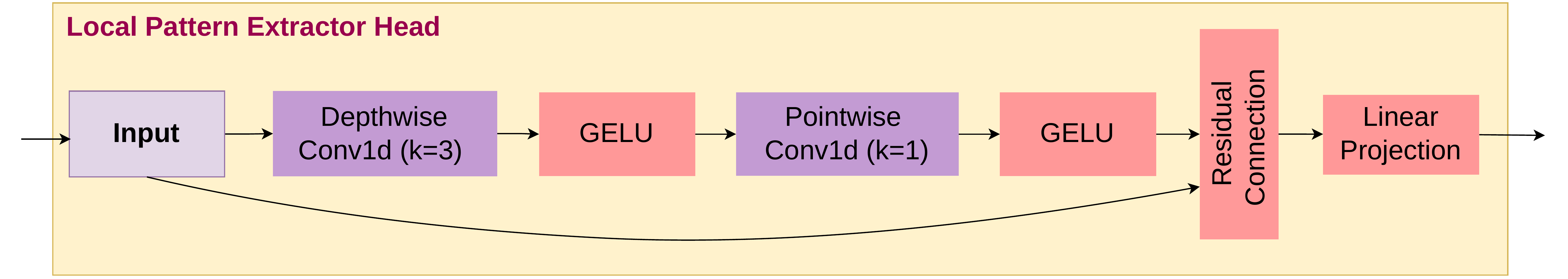}
\caption{Local Pattern Extractor Head with depthwise convolution (kernel size 3).}
\label{fig:local_head}
\end{figure}

\begin{figure}[t]
\centering
\includegraphics[width=0.99\textwidth]{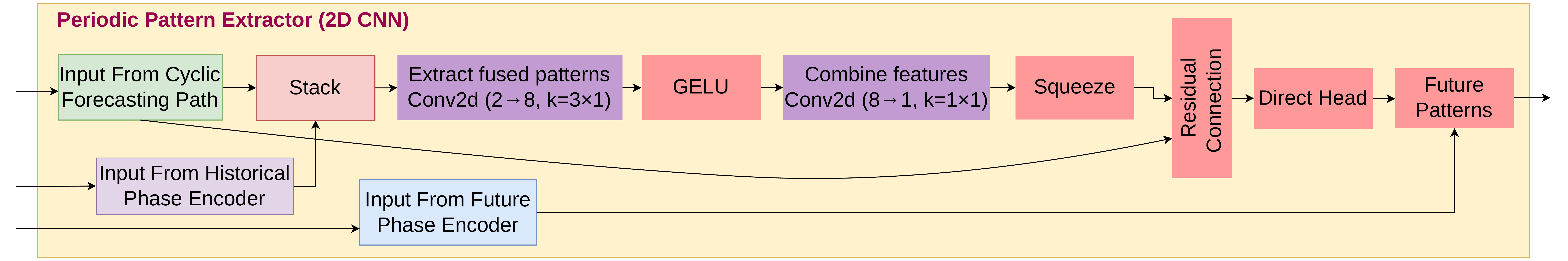}
\caption{Periodic Pattern Extractor Head using Conv2D to fuse input with cyclic past embeddings, adding cyclic future embeddings.}
\label{fig:periodic_head}
\end{figure}

The network comprises six principal modules: (i) \textit{RevIN normalization} is applied to the input, normalizing each channel independently using statistics computed from the look-back window. The normalization statistics are stored for later denormalization, enabling the model to focus on relative temporal patterns rather than absolute values. (ii) \textit{Gate feature extraction} computes a set of statistical descriptors from the normalized input, including the per-channel mean, standard deviation, last value, overall slope, and average absolute change between consecutive time steps. These compact features serve as the basis for all subsequent adaptive gating mechanisms. These gate features are normalized via LayerNorm before being passed to each 
gating MLP, ensuring that the five statistics have comparable scales. (iii) A \textit{channel mixer} enables cross-channel interaction while maintaining computational efficiency. It employs a lightweight bottleneck MLP with a residual connection, where the bottleneck dimension adapts to the number of input channels. This allows information to flow across different variates without adding significant parameters, with dropout rate 0.1 applied between the two linear layers. (iv) The \textit{Cyclic Forecasting Path} incorporates explicit seasonal inductive bias through four learnable cyclic embedding tables. For each dataset, we set a base period $P$ based on the sampling frequency (24 for hourly data, 96 for 15-minute data, 144 for 10-minute data, 7 for daily data) and derive four cyclic 
periods as $\{P/2, P, 2P, \min(7P, 1024)\}$, where the last period is capped at 1024 to avoid excessively large embedding tables. This yields periods of 
$\{12, 24, 48, 168\}$ for hourly datasets, $\{48, 96, 192, 672\}$ for 15-minute datasets, $\{72, 144, 288, 1008\}$ for 10-minute datasets, and $\{3, 7, 14, 49\}$ for daily datasets. The path combines these cyclic embeddings using learned 
period weights generated from the gate features, enabling the model to emphasize the most relevant periodicities for each input sample. The cyclic past embeddings 
$g_{\text{past}}$ are normalized via LayerNorm before being passed to the forecasting heads. Table~\ref{tab:periods} summarizes the actual periods used for each dataset.

\begin{table}[t]
\centering
\caption{Cyclic periods used for each dataset. The base period $P$ is set from 
the sampling frequency, and the four cyclic periods are derived as 
$\{P/2, P, 2P, \min(7P, 1024)\}$.}
\label{tab:periods}
\scriptsize
\renewcommand{\arraystretch}{1.2}
\setlength{\tabcolsep}{3pt}
\begin{tabular}{@{}lccc@{}}
\toprule
\textbf{Dataset} & \textbf{Sampling Interval} & \textbf{Base Period $P$} & \textbf{Cyclic Periods $\{P/2, P, 2P, 7P\}$} \\
\midrule
ETTh1 / ETTh2 & 1 hour (60 min) & 24 & \{12, 24, 48, 168\} \\
ETTm1 / ETTm2 & 15 min & 96 & \{48, 96, 192, 672\} \\
Weather & 10 min & 144 & \{72, 144, 288, 1008\} \\
ECL & 1 hour (60 min) & 24 & \{12, 24, 48, 168\} \\
Traffic & 1 hour (60 min) & 24 & \{12, 24, 48, 168\} \\
Exchange & 1 day (1440 min) & 7 & \{3, 7, 14, 49\} \\
PEMS03/04/07/08 & 5 min & 288 & \{144, 288, 576, 1024\} \\
\bottomrule
\end{tabular}
\end{table}

For datasets with high sampling frequencies (PEMS at 5 min, Weather at 10 min), 
the scaled periods exceed the lookback window $L=96$. In these cases, the cyclic 
path observes only a fraction of each period within a single window, and the 
MSTCN-CA path compensates by capturing longer-range dependencies through dilated 
convolutions. This is consistent with the ablation results in Section~\ref{sec:Ablation}, 
where removing the Cyclic path causes minimal degradation on PEMS datasets.

We now describe the second path. (v) The \textit{MSTCN-CA Path} (Fig.~\ref{fig:mstcn_ca}) captures multi-scale temporal dependencies using a lightweight Temporal Convolutional Network. The 
backbone consists of two stacked Temporal Multi-Scale (TMS) blocks, each 
employing four parallel depthwise dilated convolutions with dilation rates of 
1, 2, 4, and 8. RMSNorm is used instead of LayerNorm to preserve magnitude 
information, and channel-adaptive gating modulates the contribution of the 
multi-scale features based on the input summary. In RhyMix, multi-scale refers 
to capturing temporal patterns at different resolutions. The dilation rates 
$d \in \{1, 2, 4, 8\}$ allow the model to capture short-term fluctuations (d=1), 
medium-term trends (d=2,4), and long-term dependencies (d=8) (see 
Section~\ref{sec:mathematical} for the formal computation). This is complemented 
by the Cyclic Path, which uses dataset-specific periods $P_k$ scaled from the 
hourly base set $\{12, 24, 48, 168\}$ (see Table~\ref{tab:periods} for the 
actual values per dataset) to model half-day, daily, two-day, and weekly cycles. (vi) The \textit{multi-scale fusion module} integrates outputs from both pathways through two levels of adaptive combination. First, a path gate learns to combine four specialized forecasting heads for each sample and channel. The four heads capture different temporal patterns: (a) a direct linear projection, (b) trend-seasonal decomposition using an adaptive moving average with kernel size $k=25$ for $L=96$ and reflection padding ($\lfloor 25/2 \rfloor = 12$ on each side) to preserve sequence length and handle boundary effects, (c) local convolutional features with kernel size 3 (Fig.~\ref{fig:local_head}), and (d) periodic fusion with cyclic embeddings using a 2D CNN (Fig.~\ref{fig:periodic_head}). Both the Cyclic and MSTCN-CA Paths apply all four heads to their respective representations. We employ four heads to cover the dominant temporal patterns in time series—linear, 
trend-seasonal, local, and periodic—avoiding the limitations of a single head and the redundancy of more (detailed formulations are provided in Section~\ref{sec:mathematical}). Second, a hybrid gate (a two-layer MLP with hidden dimension 16 and ReLU activation) learns to balance the contributions of the Cyclic and MSTCN-CA Paths, producing the final fused forecast.



\subsubsection{Mathematical Modeling}
\label{sec:mathematical}

We address the problem of forecasting future values $\hat{Y}_{1:H} \in \mathbb{R}^{H \times C}$ from multivariate time series $X_{1:L} \in \mathbb{R}^{L \times C}$ with the look-back window $L=96$ and $C$ channels. The complete forward process is detailed below.

The input $X \in \mathbb{R}^{B \times L \times C}$ ($B$ is the batch size) is normalized using RevIN:
\begin{equation}
X_{\text{norm}} = \frac{X - \mu_X}{\sigma_X + \epsilon} \odot g + b
\end{equation}

where $\mu_X, \sigma_X \in \mathbb{R}^{B \times 1 \times C}$ are the per-channel mean and standard deviation, and $\odot$ denotes element-wise multiplication, whereas $g, b \in \mathbb{R}^{C}$ are learnable scale and shift parameters and $\epsilon = 10^{-5}$ is a small constant for numerical stability. To enable adaptive gating, we extract five statistical features per channel: mean, standard deviation, last value, slope (last minus first), and mean absolute change between consecutive steps. These are stacked as $\mathcal{F} \in \mathbb{R}^{B \times C \times 5}$, forming the basis for all subsequent gating mechanisms. We define $\bar{\mathcal{F}} = \text{mean}_{c}(\mathcal{F}) \in \mathbb{R}^{B \times 5}$ as the gate features averaged across channels. The stacked features $\mathcal{F}$ are normalized via LayerNorm over the 5-dimensional statistics axis (independently for each channel and batch element) before being passed to any gating MLP. Cross-channel interaction is enabled via a bottleneck MLP with residual connection:
\begin{equation}
X_{\text{mixed}} = X_{\text{norm}} + \alpha \cdot \text{MLP}(\text{LayerNorm}(X_{\text{norm}}))
\label{eq:mixer}
\end{equation}
where the MLP has structure $C \rightarrow r \rightarrow C$ with $r = \min(64, \max(8, \lfloor C/8 \rfloor))$ and $\alpha$ is a learnable scale parameter initialized to $0.05$. The LayerNorm is applied channel-wise, normalizing over the $C$ dimension at each time step. The mixed representation is processed through two complementary paths.

\emph{Cyclic Forecasting Path:} This path incorporates explicit seasonal inductive bias through four learnable cyclic embedding matrices $G_k \in \mathbb{R}^{P_k \times C}$ with periods $\{12,$ $24,$ $48,$ $168\}$. The embedding matrices are initialized with $\mathcal{N}(0, 0.02)$.

The period weights are $\lambda_k = \sigma(\text{MLP}_{\lambda}(\mathcal{F})) \in \mathbb{R}^{B \times C \times 4}$, 
where $\sigma$ denotes the sigmoid function. Unlike Softmax which forces a fixed budget of 1 across periods, Sigmoid allows the model to independently gate each 
period, enabling suppression of all periods or simultaneous activation of multiple periods. where $\phi \in \mathbb{R}^{1 \times C}$ is a phase offset initialized 
to zero and fixed during training, shared across the batch, that determines the starting position of each cycle for each channel relative to the input window. Here, $t \in \{0, \ldots, L-1\}$ indexes time steps within the look-back window, and 
$P_k \in \{12, 24, 48, 168\}$ predefined cyclic periods (half-daily, daily, bi-daily, weekly).
\begin{align}
\text{gpast} &= \frac{\sum_{k} \lambda_k G_k[(\lfloor\phi\rfloor + t) \bmod P_k]}{\sum_k \lambda_k}, \quad &
\text{gfut} &= \frac{\sum_{k} \lambda_k G_k[(\lfloor\phi\rfloor + L + t) \bmod P_k]}{\sum_k \lambda_k}
\label{eq:cyclic}
\end{align}

where $\text{gpast} \in \mathbb{R}^{B \times L \times C}$ and $\text{gfut} \in \mathbb{R}^{B \times H \times C}$ and $\lfloor\phi\rfloor$ denotes the floor of $\phi$, used as the integer phase offset for modulo indexing. This multi-period design provides inductive bias for half-day, daily, two-day, and weekly cycles.

\emph{MSTCN-CA Path:} This path captures multi-scale temporal dependencies using two stacked Temporal Multi-Scale (TMS) blocks. Each TMS block employs four parallel depthwise dilated convolutions with dilation rates $d \in \{1, 2, 4, 8\}$, where $\text{Conv1D}_d$ denotes a 1D depthwise convolution with dilation rate $d$. The computation is:
\begin{align}
Z &= \text{RMSNorm}(X_{\text{mixed}}), \tag{4} \\
Y_d &= \text{Conv1D}_d(Z), \quad d \in \{1, 2, 4, 8\}, \tag{5} \\
Y_{\text{tms}} &= \frac{1}{4}\sum_{d \in \{1, 2, 4, 8\}} \text{GELU}(Y_d), \tag{6} \\
\Gamma &= \sigma\left(\text{MLP}\left(\frac{1}{L}\sum_{t=1}^{L} X_{\text{mixed},t}\right)\right) \in \mathbb{R}^{B \times 1 \times C}, \tag{7} \\
X_{\text{tcn}} &= X_{\text{mixed}} + \beta \cdot \text{Dropout}(Y_{\text{tms}}) \odot \Gamma, \tag{8}
\end{align}
where the MLP uses ReLU activation between two linear layers, $\beta$ is a learnable scale parameter initialized to 0.10. The channel attention $\Gamma$ modulates multi-scale features based on input characteristics.

\emph{Four Forecasting Heads:} To capture the four dominant pattern types in time series—linear trends, trend-seasonal components, local fluctuations, and periodic cycles—we employ four specialized heads. A single head cannot capture all four, while more than four heads 
introduce redundancy and overfitting. Both paths—Cyclic and MSTCN-CA—apply these four heads to their respective representations. We use the symbol $Z$ to denote the input to the heads: $Z = X_{\text{mixed}}$ for the Cyclic Path and $Z = X_{\text{tcn}}$ for the MSTCN-CA Path, enabling each path to generate predictions from all four perspectives:  (i) Direct linear projection $\hat{Y}_{\text{direct}} = W_{\text{direct}} \cdot Z + b_{\text{direct}}$; (ii) Trend-Seasonal decomposition using moving average 
($\text{trend} = \frac{1}{k}\sum_{i=0}^{k-1} Z_{t-i}$, $k=25$ for $L=96$), 
$\text{seasonal} = Z - \text{trend}$, 
$\hat{Y}_{\text{decomp}} = W_{\text{seasonal}} \cdot \text{seasonal} + W_{\text{trend}} \cdot \text{trend}$; (iii) Local convolution with kernel size 3: $Z_{\text{local}} = \text{Conv1D}(Z) + Z$, $\hat{Y}_{\text{local}} = W_{\text{local}} \cdot Z_{\text{local}}$; and (iv) Periodic fusion using Conv2D: 
$Z_{\text{periodic}} = \text{Conv2D}([Z, \text{gpast}]) + Z$, $\hat{Y}_{\text{periodic}} = W_{\text{periodic}} \cdot Z_{\text{periodic}} + \text{gfut}$, 
where $[Z, \text{gpast}]$ denotes channel-wise concatenation and $\text{gfut}$ injects explicit seasonal information into the forecast. The path weights combine these heads: $\hat{Y}_1 = \hat{Y}_{\text{direct}}$, 
$\hat{Y}_2 = \hat{Y}_{\text{decomp}}$, 
$\hat{Y}_3 = \hat{Y}_{\text{local}}$, 
$\hat{Y}_4 = \hat{Y}_{\text{periodic}}$. The path gate produces per-channel head weights $w = \text{Softmax}(\text{MLP}_{w}(\mathcal{F})) \in \mathbb{R}^{B \times C \times 4}$, where $\mathcal{F}$ is the full statistics tensor (not averaged across channels). This allows each channel to emphasize different forecasting heads based on its own temporal statistics. The cyclic period weights $\lambda_k \in \mathbb{R}^{B \times C \times 4}$ are also computed per-channel, while the hybrid gate $h = \text{Softmax}(\text{MLP}_h(\bar{\mathcal{F}})) \in \mathbb{R}^{B \times 2}$ operates on the channel-averaged features $\bar{\mathcal{F}}$.
\label{eq:pathgate}

The output of each path is $\hat{Y}_{\text{path}} = \sum_{i=1}^{4} w_i \odot \hat{Y}_i$, 
where $w_i \in \mathbb{R}^{B \times C}$ is the $i$-th head weight for each channel. The Cyclic Path ($\hat{Y}_{\text{cyclic}}$) and MSTCN-CA Path ($\hat{Y}_{\text{tcn}}$) outputs are fused via a hybrid gate:
\begin{equation}
h = \text{Softmax}(\text{MLP}_{h}(\bar{\mathcal{F}})) \in \mathbb{R}^{B \times 2}
\label{eq:hybrid}
\end{equation}
The final forecast is $\hat{Y} = h_0 \odot \hat{Y}_{\text{cyclic}} + h_1 \odot \hat{Y}_{\text{tcn}} \in \mathbb{R}^{B \times H \times C}$, adaptively balancing seasonal and data-driven features. The final prediction is denormalized using RevIN:
\begin{equation}
\hat{Y}_{\text{final}} = \frac{\hat{Y} - b}{g + \epsilon} \odot (\sigma_X + \epsilon) + \mu_X
\label{eq:denorm}
\end{equation}
The model is optimized using Huber loss with $\delta=1.0$. The Huber loss behaves like mean squared error for small errors and like mean absolute error for large errors, making it less sensitive to outliers while remaining smooth near the optimum:
\begin{equation}
\mathcal{L}_{\delta}(a) = \begin{cases}
0.5a^2 & |a| \leq \delta \\
\delta(|a| - 0.5\delta) & \text{otherwise}
\end{cases}
\label{eq:huber}
\end{equation}
The loss is averaged over batch, horizon, and channel dimensions (mean reduction). where $a = \hat{Y}_{\text{final}} - Y_{\text{true}}$. The total loss is:
\begin{equation}
\mathcal{L}_{\text{total}} = \frac{1}{BHC}\sum_{i=1}^{B}\sum_{t=1}^{H}\sum_{c=1}^{C} \mathcal{L}_{\delta}(\hat{Y}_{i,t,c}, Y_{i,t,c})
\label{eq:tcn_total_loss}
\end{equation}

\subsection{How is RhyMix Different from Prior Works?}

    \begin{table}[!htb]
    \centering
    \scriptsize
    \caption{Taxonomy of model architectures by Channel-Interaction strategy. Different state-of-the-art models align with the established Channel Interaction (CI, CD, CP) strategies. (Asym.: Asymmetry, Lag.: Lagginess, Pol.: Polarity, Gw.: group-wise, Dyn.: Dynamism, Ms.: Multi-scale).}
    \label{tab:channel} 
    
    \setlength{\tabcolsep}{3pt}
    
    \begin{tabular}{@{}l p{5cm} | c c c c c c | l@{}}
    \toprule
    \textbf{Strategy} & \textbf{Mechanism} &
    \multicolumn{6}{c|}{\textbf{Characteristic}} &
    \textbf{Method} \\
    \cmidrule(lr){3-8}
    & & \textbf{Asym.} & \textbf{Lag.} & \textbf{Pol.} & \textbf{Gw.} & \textbf{Dyn.} & \textbf{Ms.} & \\
    \midrule
    \textbf{CI} & - & - & - & - & - & - & - & PatchTST \\
    CI & - & - & - & - & - & - & - & CycleNet \\
    CI & - & - & - & - & - & - & - & DLinear \\
    CI & - & - & - & - & - & - & - & Timer \\
    CI & - & - & - & - & - & - & - & Chronos \\
    CI & - & - & - & - & - & - & - & LLM4TS \\
    CI & - & - & - & - & - & - & - & Time-LLM \\
    CI & - & - & - & - & - & - & - & RevIN \\
    \midrule
    \textbf{CD} & CNN-based & $\checkmark$ & - & - & - & - & - & Informer \\
    CD & CNN-based & $\checkmark$ & - & - & - & - & - & Autoformer \\
    CD & CNN-based & $\checkmark$ & - & - & - & - & - & FEDformer \\
    CD & CNN-based & $\checkmark$ & - & - & - & - & - & TimesNet \\
    CD & MLP-based & $\checkmark$ & - & - & - & - & - & TSMixer \\
    CD & MLP-based & $\checkmark$ & - & - & - & - & - & TTM \\
    CD & Transformer-based & $\checkmark$ & - & - & - & $\checkmark$ & - & iTransformer \\
    CD & Transformer-based & $\checkmark$ & - & - & - & $\checkmark$ & - & Crossformer \\
    CD & Transformer-based & $\checkmark$ & $\checkmark$ & - & - & - & - & VCformer \\
    CD & Transformer-based & $\checkmark$ & $\checkmark$ & - & - & $\checkmark$ & - & MOIRAI \\
    CD & Transformer-based & $\checkmark$ & - & - & - & $\checkmark$ & - & UniTS \\
    CD & GNN-based & - & - & - & $\checkmark$ & - & - & GTS \\
    CD & GNN-based & $\checkmark$ & - & - & - & - & $\checkmark$ & MSGNet \\
    CD & GNN-based & - & $\checkmark$ & - & - & - & - & FourierGNN \\
    CD & GNN-based & - & $\checkmark$ & - & - & - & - & FC-STGNN \\
    CD & GNN-based & $\checkmark$ & - & - & - & $\checkmark$ & - & TPGNN \\
    CD & Linear Channel Mixing & $\checkmark$ & - & - & - & - & - & SOFTS \\
    CD & Others & $\checkmark$ & - & - & - & $\checkmark$ & - & C-LoRA \\
    \midrule
    \textbf{CP} & CNN-based & $\checkmark$ & - & - & - & - & - & ModernTCN \\
    CP & Transformer-based & $\checkmark$ & - & - & $\checkmark$ & - & - & DUET \\
    CP & Transformer-based & $\checkmark$ & - & - & - & $\checkmark$ & - & MCformer \\
    CP & Transformer-based & $\checkmark$ & - & - & $\checkmark$ & $\checkmark$ & - & DGCformer \\
    CP & Transformer-based & - & - & - & - & $\checkmark$ & - & CM \\
    CP & GNN-based & $\checkmark$ & - & - & - & - & - & MTGNN \\
    CP & GNN-based & $\checkmark$ & - & $\checkmark$ & - & - & - & CrossGNN \\
    CP & GNN-based & $\checkmark$ & - & - & $\checkmark$ & - & - & WaveForM \\
    CP & GNN-based & - & - & - & - & $\checkmark$ & - & MTSF-DG \\
    CP & GNN-based & $\checkmark$ & - & - & $\checkmark$ & - & - & ReMo \\
    CP & GNN-based & $\checkmark$ & - & - & $\checkmark$ & $\checkmark$ & $\checkmark$ & Ada-MSHyper \\
    CP & Others & $\checkmark$ & $\checkmark$ & - & - & - & - & LIFT \\
    CP & Others & $\checkmark$ & - & - & $\checkmark$ & - & - & CCM \\

    \midrule
    \textbf{CI + CD} & Parallel CNN (CI) and BiLSTM/Transformer(CD) based Dual Encoder & $\checkmark$ & - & - & - & $\checkmark$ & $\checkmark$ & MSTN \\
    \midrule
    \textbf{CI + CD} & \textbf{Parallel Cyclic Forecasting Path (CI) and MSTCN-CA Contextual Forecasting (CI) + Channel Mixer (CD)} & $\checkmark$ & $\checkmark$ & $\checkmark$ & $\checkmark$ & $\checkmark$ & $\checkmark$ & \textbf{RhyMix} \\
    
    \bottomrule
    \end{tabular}
    \end{table}


The established field of multivariate time series forecasting is conventionally organized around three channel interaction strategies--CI, CD, and CP approaches~\citep{qiu2025comprehensivesurveydeeplearning, shevtekar2026mstn}--as summarized in Table~\ref{tab:channel}. These strategies specify how a model encodes and exploits inter-variable dependencies across the $C$ channels of a multivariate signal.

\textbf{CI:} CI methods (e.g., DLinear, PatchTST, CycleNet) process each variable independently using lightweight temporal operators such as linear layers or multilayer perceptrons. Their advantage lies in linear channel complexity $\mathcal{O}(C)$ and the efficient temporal complexity $\mathcal{O}(L)$, although this comes at the expense of eliminating the cross-channel structure.
    
 \textbf{CD:} CD approaches (e.g., iTransformer, Crossformer, TimesNet) explicitly model inter-variable correlation through dense attention, convolution, or graph-based mechanisms. While these models offer comprehensive feature coupling, they incur high computational cost--typically $\mathcal{O}(C^2)$ or $\mathcal{O}(L^2)$---which limits their scalability and latency characteristics.
    
\textbf{CP:} CP models (e.g. ModernTCN, MCformer, MTGNN) map the channel dimension into a low-rank latent space before applying interaction, thus avoiding the full $\mathcal{O}(C^2)$ cost of CD models. Although channel-efficient, these models often impose simplified temporal assumptions, reducing their ability to capture multi-scale dynamics.
    
\textbf{CI + CD:} By combining the complementary strengths of CI and CD strategies with explicit multi-period cyclic priors and adaptive gating, RhyMix satisfies six key architectural criteria—Asymmetry ($L$ vs $H$ embeddings), Lagginess (dilated convs $d=1,2,4,8$), Polarity (trend-seasonal decomposition), Group-wise (channel mixer bottleneck $C \rightarrow r \rightarrow C$), Dynamism (adaptive gates $\lambda_k$, $w$, $h$), and Multi-scale (dilations $+$ periods). This comprehensive design enables RhyMix to deliver improved long-term forecasting performance while remaining lightweight and efficient.

\textbf{Positioning.} We emphasize that the individual components of RhyMix — 
cyclic embeddings, multi-scale dilated convolutions, channel attention, and 
gating mechanisms — are each established in prior work. The methodological 
contribution of this paper is not any single component, but the 
\emph{two-level adaptive gating framework} that combines them: a per-channel  path gate that fuses four specialized forecasting heads, and a per-sample  hybrid gate that fuses the Cyclic and MSTCN-CA paths. Both gates are driven by 
the same input statistics tensor $\mathcal{F}$, and together they allow the  model to adapt its behavior at both the head level and the path level based on  each sample's temporal characteristics. No prior work combines input-statistics-driven gating at both levels. CycleNet uses a single fixed period; ModernTCN 
lacks seasonal priors; MSTN uses fixed-weight fusion between its two encoders. 
RhyMix's two-level adaptive gating is, to our knowledge, a new design for multivariate forecasting.

    \begin{table*}[t]
\centering
\scriptsize
\setlength{\tabcolsep}{0.5pt} 
\caption{Comparison of architectural characteristics. Evaluation of RhyMix against CI, CD, and CP channel interaction strategies.}
\label{tab:strategy_comparison}
\begin{tabular}{@{}p{2.6cm}p{0.8cm}p{1.0cm}p{1.1cm}p{1.1cm}p{5.5cm}@{}}
\toprule
\textbf{Dimension} & \textbf{CI} & \textbf{CD} & \textbf{CP} & \textbf{RhyMix} & \textbf{Rationale} \\
\midrule
\textbf{Efficiency} & High & Low & Moderate & High & Linear complexity $\mathcal{O}(L \cdot C)$; depthwise convolutions; bottleneck channel mixer. \\
\textbf{Parameter Count} & Low & High & Moderate & Low & $\sim$40K params ($C=7$, $H=96$). \\
\textbf{Generalizability} & Low & Moderate & High & High & Adaptive gating for dynamic input adjustment. \\
\textbf{Capacity} & Low & High & Moderate & High & MSTCN-CA (multi-scale); Cyclic (seasonal). \\
\textbf{Multi-Scale} & Low & Moderate & Low & High & Dilations $d\in\{1,2,4,8\}$ for multi-scale. \\
\textbf{Seasonal Priors} & No & No & No & Yes & Cyclic tables with periods 12, 24, 48, 168. \\
\bottomrule
\end{tabular}

\end{table*}

    Table~\ref{tab:strategy_comparison} summarizes how RhyMix combines desirable characteristics from different design approaches through its hybrid architecture. The model achieves high capacity through its dual-path design while maintaining efficiency through linear complexity and low parameter count.

    \section{Experiments}
    \label{sec4}
    
    This section presents a comprehensive evaluation of RhyMix on long-term forecasting tasks using twelve publicly available benchmark datasets, covering diverse domains including electricity, traffic, weather, and finance.

\subsection{Benchmark Datasets}
    \label{Benchmarkdata}

\begin{table}
\centering
\caption{Benchmark Datasets for Long-Term Forecasting.}
\label{tab:dataset_overview}
\scriptsize
\renewcommand{\arraystretch}{1.2}
\setlength{\tabcolsep}{2pt} 
\begin{tabular}{@{} l l c c c p{3.2cm} @{}}
\hline
\textbf{Task} & \textbf{Datasets} & \textbf{Features} & \textbf{Horizons} & \textbf{Dataset Size} & \textbf{Information} \\
\hline
\textbf{Forecasting} & ETTh1 / ETTh2 & 7 & \{96, 192, 336, 720\} & (8,545, 2,881, 2,881) & Electricity (Hourly) \\
 & ETTm1 / ETTm2 & 7 & \{96, 192, 336, 720\} & (34,465, 11,521, 11,521) & Electricity (15 mins) \\
 & Electricity (ECL) & 321 & \{96, 192, 336, 720\} & (18,317, 2,633, 5,261) & Electricity (Hourly) \\
 & Traffic & 862 & \{96, 192, 336, 720\} & (12,185, 1,757, 3,509) & Transportation (Hourly) \\
 & Weather & 21 & \{96, 192, 336, 720\} & (36,792, 5,271, 10,540) & Weather (10 mins) \\
 & Exchange & 8 & \{96, 192, 336, 720\} & (5,120, 665, 1,422) & Exchange Rate (Daily) \\
 & PEMS03 & 358 & \{12, 24, 48, 96\} & (15,724, 5,241, 5,243) & Transportation (5 mins) \\
 & PEMS04 & 307 & \{12, 24, 48, 96\} & (10,195, 3,398, 3,399) & Transportation (5 mins) \\
 & PEMS07 & 883 & \{12, 24, 48, 96\} & (16,934, 5,644, 5,646) & Transportation (5 mins) \\
 & PEMS08 & 170 & \{12, 24, 48, 96\} & (10,713, 3,571, 3,572) & Transportation (5 mins) \\
\hline
\end{tabular}
\end{table}

    We evaluate RhyMix on twelve publicly available benchmark datasets (Table~\ref{tab:dataset_overview}). These include four ETT datasets (ETTh1, ETTh2, ETTm1, ETTm2) for electricity transformer temperature, Electricity ECL (321 clients), Traffic (862 sensors), Weather (21 variables), Exchange (8 currencies), and four PEMS traffic datasets (PEMS03, PEMS04, PEMS07, PEMS08). All datasets follow standard protocols~\citep{liu2024itransformerinvertedtransformerseffective, nie2023timeseriesworth64, wu2023timesnettemporal2dvariationmodeling} and are publicly available~\citep{lin_california_traffic_2025, liu2022scinettimeseriesmodeling, 877p-as67_2025, wu2022autoformer}.

    \subsection{Baselines} 
\label{sec:baselines}

For the ETT, ECL, Traffic, Weather, and Exchange datasets, we compare RhyMix with latest baselines including Time-o1~\citep{wang2026timeo}, TimeMixer++~\citep{wang2025timemixergeneraltimeseries}, GCMNet~\citep{LIU2026113287}, SEG-MOE~\citep{ortigossa2026segmoemultiresolutionsegmentwisemixtureofexperts}, SOFTS~\citep{han2024softsefficientmultivariatetime}, FilterNet (TexFilter)~\citep{yi2024filternet}, and PatchTST~\citep{nie2023timeseriesworth64}.

For PEMS traffic datasets, we compare with MSTN-Transformer~\citep{shevtekar2026mstn}, MSTN-BiLSTM~\citep{shevtekar2026mstn}, iTransformer~\citep{liu2024itransformerinvertedtransformerseffective}, ModernTCN~\citep{donghao2024moderntcn}, PatchTST~\citep{nie2023timeseriesworth64}, TSMixer~\citep{chen2023tsmixer}, and Autoformer~\citep{wu2022autoformer}. The baseline sets differ between the main 8 datasets and the PEMS 4 datasets. MSTN~\citep{shevtekar2026mstn} was proposed for traffic forecasting and reports results only on PEMS benchmarks; we include it where its results are available. Second, SEG-MOE~\citep{ortigossa2026segmoemultiresolutionsegmentwisemixtureofexperts} and GCMNet~\citep{LIU2026113287} do not report the PEMS results.

    \subsection{Experimental Setup}
\label{exp_setup}

\subsubsection{Data Preprocessing and Evaluation Protocol}
\label{sec:data_protocol}

Our preprocessing and splits follow established practices~\citep{liu2024itransformerinvertedtransformerseffective, wu2023timesnettemporal2dvariationmodeling, shevtekar2026mstn, nie2023timeseriesworth64}. All datasets are split chronologically to preserve temporal causality. For ETT datasets, we use a 60\%/20\%/20\% train/validation/test split. For ECL, Traffic, Weather, and Exchange, we adopt a 70\%/10\%/20\% split. For PEMS datasets (PEMS03, PEMS04, PEMS07, PEMS08), we use a 60\%/20\%/20\% split following iTransformer~\citep{liu2024itransformerinvertedtransformerseffective}. To prevent information leakage, we partition raw time series into train/validation/test sets and construct sliding windows within each split independently. Normalization is performed using statistics computed exclusively from training data, which are then applied to validation and test sets. Following the standard benchmark protocol~\citep{wu2023timesnettemporal2dvariationmodeling, liu2024itransformerinvertedtransformerseffective}, we use a fixed look-back window of $L=96$ across all datasets. Prediction horizons are $H \in \{96, 192, 336, 720\}$ for ETT, ECL, Traffic, Weather, and Exchange; and $H \in \{12, 24, 48, 96\}$ for PEMS datasets. Metrics (MSE and MAE) are reported on the normalized scale, consistent with prior benchmarks.

\subsubsection{Training and Model Configuration}
\label{modeltraining}

RhyMix is implemented in Python 3.13.1 using PyTorch 2.7.1 and trained on an NVIDIA DGX A100 server (8 $\times$ NVIDIA A100-SXM4 GPUs, 40 GB VRAM each). The architecture consists of a Cyclic Forecasting Path (periods 12, 24, 48, 168) and an MSTCN-CA Contextual Forecasting Path (two TMS blocks, dilations 1, 2, 4, 8). Both paths share four forecasting heads: Direct, Trend-Seasonal decomposition, Local convolution, and Periodic fusion. Adaptive gating mechanisms dynamically combine the outputs of both paths. Training uses the AdamW optimizer with learning rate $7 \times 10^{-4}$ and weight decay $1 \times 10^{-4}$, with gradient clipping of 1.0. The learning 
rate is annealed using a cosine schedule from $7 \times 10^{-4}$ to $1 \times 10^{-6}$ 
over the training epochs. Training runs for a maximum of 50 epochs, with early stopping of patience 6 based on validation MSE. The Huber loss ($\delta=1.0$) provides robustness to outliers. All experiments use $L=96$. Horizons are $H \in \{96, 192, 336, 720\}$ (ETT, ECL, Traffic, Weather, Exchange) and $H \in \{12, 24, 48, 96\}$ (PEMS). Hyperparameters follow standard practice. All experiments are repeated with 5 random seeds; results are reported as mean $\pm$ std.

\subsubsection{Evaluation Metrics}

We evaluate long-term forecasting performance using Mean Squared Error (MSE) and Mean Absolute Error (MAE), following the standard protocols of iTransformer~\citep{liu2024itransformerinvertedtransformerseffective}, TimesNet~\citep{wu2023timesnettemporal2dvariationmodeling}, SOFTS~\citep{han2024softsefficientmultivariatetime}, and MSTN~\citep{shevtekar2026mstn}.

\subsection{Long-Term Forecasting Results}

\begin{table*}
\centering
\scriptsize
\caption{Multivariate long-term forecasting results. Input look-back window $L=96$. Prediction horizons $H \in \{96, 192, 336, 720\}$. Avg is averaged from all four prediction lengths. \textcolor{red}{\textbf{Red}}: 1st rank, \textcolor{blue}{\textbf{Blue}}: 2nd rank. We evaluate each model five times and report the mean (standard deviation); results marked with \underline{underline} are not significantly worse than the first rank (Wilcoxon signed-rank test with Holm–Bonferroni correction, $\alpha = 0.05$). Example: 0.010(0.001) means 0.010 $\pm$ 0.001. TimeM.: TimeMixer++, Impr.: Improvement \%.}
\label{tab:forecasting_comparison}
\setlength{\tabcolsep}{0.4pt}
\begin{tabular}{@{}l*{8}{c}@{}}
\toprule
\textbf{Models} & \textbf{RhyMix} & \textbf{TimeM.++} & \textbf{Time-o1} & \textbf{GCMNet} & \textbf{SEG-MOE} & \textbf{SOFTS} & \textbf{TexFilter} & \textbf{PatchTST} \\
 
\textbf{Metric} & \textbf{MSE MAE} & \textbf{MSE MAE} & \textbf{MSE MAE} & \textbf{MSE MAE} & \textbf{MSE MAE} & \textbf{MSE MAE} & \textbf{MSE MAE} & \textbf{MSE MAE} \\
\midrule

\textbf{ETTh1} & & & & & & & & \\
 96 & \textcolor{red}{\textbf{0.360(0.01)}} \textcolor{red}{\textbf{0.382(0.01)}} & \textcolor{blue}{\textbf{\underline{0.361}}} 0.403 & \underline{0.368} \textcolor{blue}{\textbf{\underline{0.391}}} & 0.390 0.406 & 0.370 0.401 & 0.381 \underline{0.399} & 0.382 0.402 & 0.394 0.406 \\
 192 & \textcolor{red}{\textbf{0.415(0.01)}} \textcolor{red}{\textbf{0.411(0.01)}} & \textcolor{blue}{\textbf{\underline{0.416}}} 0.441 & \underline{0.424} \textcolor{blue}{\textbf{\underline{0.422}}} & 0.438 0.434 & 0.435 0.430 & 0.435 0.431 & 0.430 0.429 & 0.440 0.435 \\
 336 & \textcolor{blue}{\textbf{0.449(0.02)}} \textcolor{red}{\textbf{0.429(0.01)}} & \textcolor{red}{\textbf{0.430}} \textcolor{blue}{\textbf{\underline{0.434}}} & 0.467 0.441 & 0.474 0.454 & 0.507 0.456 & 0.480 0.452 & 0.472 0.451 & 0.491 0.462 \\
 720 & \textcolor{red}{\textbf{0.459(0.02)}} \textcolor{red}{\textbf{0.445(0.02)}} & \underline{0.467} \textcolor{blue}{\textbf{\underline{0.451}}} & \textcolor{blue}{\textbf{\underline{0.465}}} 0.463 & 0.477 0.478 & 0.579 0.485 & 0.499 0.488 & 0.481 0.473 & 0.487 0.479 \\
 Avg & \textcolor{blue}{\textbf{0.421}} \textcolor{red}{\textbf{0.417}} & \textcolor{red}{\textbf{0.419}} 0.432 & 0.431 \textcolor{blue}{\textbf{0.429}} & 0.445 0.443 & 0.473 0.441 & 0.449 0.442 & 0.441 0.439 & 0.453 0.446 \\
 \midrule
 
\textbf{ETTh2} & & & & & & & & \\
 96 & \textcolor{red}{\textbf{0.269(0.00)}} \textcolor{red}{\textbf{0.319(0.00)}} & \textcolor{blue}{\textbf{\underline{0.276}}} \textcolor{blue}{\textbf{\underline{0.328}}} & 0.282 0.330 & 0.290 0.341 & 0.307 0.347 & 0.297 0.347 & 0.293 0.343 & 0.288 0.340 \\
 192 & \textcolor{blue}{\textbf{0.349(0.01)}} \textcolor{red}{\textbf{0.369(0.01)}} & \textcolor{red}{\textbf{0.342}} \textcolor{blue}{\textbf{\underline{0.379}}} & 0.359 \underline{0.381} & 0.375 0.395 & 0.390 0.402 & 0.373 0.394 & 0.374 0.396 & 0.376 0.395 \\
 336 & 0.397(0.02) \textcolor{blue}{\textbf{0.411(0.02)}} & \textcolor{red}{\textbf{0.346}} \textcolor{red}{\textbf{0.398}} & \textcolor{blue}{\textbf{0.394}} 0.414 & 0.414 0.428 & 0.434 0.441 & 0.410 0.426 & 0.417 0.430 & 0.440 0.451 \\
 720 & 0.410(0.02) 0.429(0.02) & \textcolor{red}{\textbf{0.392}} \textcolor{red}{\textbf{0.415}} & \textcolor{blue}{\textbf{0.400}} \textcolor{blue}{\textbf{0.427}} & 0.423 0.440 & 0.474 0.474 & 0.411 0.433 & 0.449 0.460 & 0.436 0.453 \\
 Avg & \textcolor{blue}{\textbf{0.356}} \textcolor{blue}{\textbf{0.382}} & \textcolor{red}{\textbf{0.339}} \textcolor{red}{\textbf{0.380}} & 0.359 0.388 & 0.376 0.401 & 0.401 0.416 & 0.373 0.400 & 0.383 0.407 & 0.385 0.410 \\
 \midrule
 
\textbf{ETTm1} & & & & & & & & \\
 96 & \textcolor{red}{\textbf{0.302(0.00)}} \textcolor{blue}{\textbf{0.339(0.00)}} & \textcolor{blue}{\textbf{\underline{0.310}}} \textcolor{red}{\textbf{0.334}} & 0.321 0.357 & 0.330 0.370 & 0.323 0.363 & 0.325 0.361 & 0.321 0.361 & 0.329 0.365 \\
 192 & \textcolor{blue}{\textbf{0.352(0.01)}} \textcolor{blue}{\textbf{0.369(0.01)}} & \textcolor{red}{\textbf{0.348}} \textcolor{red}{\textbf{0.362}} & \underline{0.360} 0.378 & 0.369 0.389 & 0.392 0.395 & 0.375 0.389 & \underline{0.367} 0.387 & 0.380 0.394 \\
 336 & \textcolor{blue}{\textbf{0.381(0.02)}} \textcolor{red}{\textbf{0.390(0.02)}} & \textcolor{red}{\textbf{0.376}} \textcolor{blue}{\textbf{\underline{0.391}}} & 0.389 \textcolor{blue}{\textbf{\underline{0.400}}} & 0.398 0.410 & 0.455 0.431 & 0.405 0.412 & 0.401 0.409 & 0.400 0.410 \\
 720 & \textcolor{red}{\textbf{0.429(0.02)}} \textcolor{red}{\textbf{0.419(0.02)}} & \textcolor{blue}{\textbf{\underline{0.440}}} \textcolor{blue}{\textbf{\underline{0.423}}} & 0.447 0.435 & 0.457 0.442 & 0.583 0.486 & 0.466 0.447 & 0.477 0.448 & 0.475 0.453 \\
 Avg & \textcolor{red}{\textbf{0.366}} \textcolor{blue}{\textbf{0.380}} & \textcolor{blue}{\textbf{0.369}} \textcolor{red}{\textbf{0.378}} & 0.379 0.393 & 0.389 0.403 & 0.438 0.416 & 0.393 0.403 & 0.392 0.401 & 0.396 0.406 \\
 \midrule
 
\textbf{ETTm2} & & & & & & & & \\
 96 & \textcolor{red}{\textbf{0.161(0.00)}} \textcolor{blue}{\textbf{0.244(0.00)}} & \textcolor{blue}{\textbf{\underline{0.170}}} 0.245 & 0.172 0.251 & 0.170 0.254 & 0.191 0.273 & 0.180 \textcolor{red}{\textbf{0.233}} & 0.175 0.258 & 0.184 0.264 \\
 192 & \textcolor{red}{\textbf{0.226(0.01)}} \textcolor{red}{\textbf{0.287(0.01)}} & \textcolor{blue}{\textbf{\underline{0.229}}} \textcolor{blue}{\textbf{\underline{0.291}}} & 0.235 0.294 & 0.236 0.296 & 0.266 0.323 & 0.246 0.306 & 0.240 0.301 & 0.246 0.306 \\
 336 & \textcolor{red}{\textbf{0.281(0.02)}} \textcolor{red}{\textbf{0.327(0.02)}} & 0.303 0.343 & \textcolor{blue}{\textbf{\underline{0.293}}} \textcolor{blue}{\textbf{\underline{0.333}}} & 0.296 0.335 & 0.337 0.367 & 0.319 0.352 & 0.311 0.347 & 0.308 0.346 \\
 720 & \textcolor{blue}{\textbf{0.378(0.02)}} \textcolor{red}{\textbf{0.381(0.02)}} & \textcolor{red}{\textbf{0.373}} 0.399 & 0.388 \textcolor{blue}{\textbf{\underline{0.389}}} & 0.396 0.392 & 0.453 0.431 & 0.405 0.401 & 0.414 0.405 & 0.409 0.402 \\
 Avg & \textcolor{red}{\textbf{0.262}} \textcolor{red}{\textbf{0.310}} & \textcolor{blue}{\textbf{\underline{0.269}}} \textcolor{blue}{\textbf{\underline{0.320}}} & 0.272 0.317 & 0.274 0.319 & 0.312 0.348 & 0.287 0.330 & 0.285 0.328 & 0.287 0.330 \\
 \midrule
 
\textbf{ECL} & & & & & & & & \\
 96 & \textcolor{red}{\textbf{0.134(0.01)}} \textcolor{red}{\textbf{0.221(0.01)}} & \textcolor{blue}{\textbf{\underline{0.135}}} \textcolor{blue}{\textbf{0.222}} & 0.145 0.235 & 0.142 0.241 & 0.199 0.274 & \underline{0.143} \underline{0.233} & 0.147 0.245 & 0.164 0.251 \\
 192 & \textcolor{red}{\textbf{0.145(0.01)}} \textcolor{blue}{\textbf{0.243(0.01)}} & \textcolor{blue}{\textbf{\underline{0.147}}} \textcolor{red}{\textbf{0.235}} & 0.159 0.249 & 0.159 0.254 & 0.223 0.298 & 0.158 0.248 & 0.160 0.250 & 0.173 0.262 \\
 336 & \textcolor{red}{\textbf{0.160(0.01)}} \textcolor{blue}{\textbf{0.251(0.02)}} & \textcolor{blue}{\textbf{\underline{0.164}}} \textcolor{red}{\textbf{0.245}} & 0.173 0.264 & 0.176 0.270 & 0.254 0.329 & 0.178 0.269 & 0.173 0.267 & 0.190 0.279 \\
 720 & \textcolor{red}{\textbf{0.197(0.02)}} \textcolor{red}{\textbf{0.287(0.02)}} & 0.212 0.310 & \textcolor{blue}{\textbf{\underline{0.203}}} \textcolor{blue}{\textbf{\underline{0.292}}} & 0.203 0.294 & \underline{0.326} 0.390 & 0.218 0.305 & 0.210 0.309 & 0.230 0.313 \\
 Avg & \textcolor{red}{\textbf{0.160}} \textcolor{red}{\textbf{0.251}} & \textcolor{blue}{\textbf{0.165}} \textcolor{blue}{\textbf{0.253}} & 0.170 0.260 & 0.170 0.265 & 0.250 0.323 & 0.174 0.264 & 0.173 0.268 & 0.189 0.276 \\
 \midrule
 
\textbf{Traffic} & & & & & & & & \\
 96 & 0.438(0.00) 0.267(0.00) & \textcolor{blue}{\textbf{0.392}} \textcolor{blue}{\textbf{0.253}} & 0.393 0.265 & 0.405 0.263 & 0.503 0.322 & \textcolor{red}{\textbf{0.376}} \textcolor{red}{\textbf{0.251}} & 0.430 0.294 & 0.427 0.272 \\
 192 & 0.462(0.01) 0.283(0.01) & \textcolor{blue}{\textbf{0.402}} \textcolor{red}{\textbf{0.258}} & 0.410 0.275 & 0.424 0.271 & 0.501 0.317 & \textcolor{red}{\textbf{0.398}} \textcolor{blue}{\textbf{0.261}} & 0.452 0.307 & 0.454 0.289 \\
 336 & 0.492(0.02) 0.306(0.02) & 0.428 \textcolor{red}{\textbf{0.263}} & \textcolor{blue}{\textbf{0.421}} 0.280 & 0.449 0.283 & 0.517 0.316 & \textcolor{red}{\textbf{0.415}} \textcolor{blue}{\textbf{0.269}} & 0.470 0.316 & 0.450 0.282 \\
 720 & 0.551(0.02) 0.323(0.02) & \textcolor{red}{\textbf{0.441}} \textcolor{red}{\textbf{0.282}} & 0.451 0.298 & 0.511 0.307 & 0.602 0.364 & \textcolor{blue}{\textbf{0.447}} \textcolor{blue}{\textbf{0.287}} & 0.498 0.323 & 0.484 0.301 \\
 Avg & 0.486 0.295 & \textcolor{blue}{\textbf{0.416}} \textcolor{red}{\textbf{0.264}} & 0.419 0.280 & 0.447 0.281 & 0.512 0.313 & \textcolor{red}{\textbf{0.409}} \textcolor{blue}{\textbf{0.267}} & 0.463 0.310 & 0.454 0.286 \\
 \midrule
 
\textbf{Weather} & & & & & & & & \\
 96 & \textcolor{red}{\textbf{0.142(0.01)}} \textcolor{red}{\textbf{0.189(0.01)}} & \textcolor{blue}{\textbf{0.155}} \textcolor{blue}{\textbf{0.205}} & 0.169 0.219 & 0.165 0.216 & 0.178 0.213 & 0.166 0.208 & 0.162 0.207 & 0.176 0.217 \\
 192 & \textcolor{red}{\textbf{0.194(0.01)}} \textcolor{red}{\textbf{0.241(0.01)}} & \textcolor{blue}{\textbf{\underline{0.201}}} \textcolor{blue}{\textbf{\underline{0.245}}} & 0.210 0.258 & 0.212 0.255 & 0.227 0.256 & 0.217 0.253 & 0.210 0.250 & 0.221 0.256 \\
 336 & \textcolor{blue}{\textbf{0.243(0.02)}} \textcolor{blue}{\textbf{0.272(0.02)}} & \textcolor{red}{\textbf{0.237}} \textcolor{red}{\textbf{0.265}} & 0.259 0.297 & 0.271 0.297 & 0.285 0.300 & 0.282 0.300 & 0.266 0.290 & 0.275 0.296 \\
 720 & 0.321(0.02) \textcolor{red}{\textbf{0.329(0.02)}} & \textcolor{red}{\textbf{0.312}} \textcolor{blue}{\textbf{\underline{0.334}}} & 0.327 0.349 & 0.345 0.346 & 0.373 0.361 & 0.356 0.351 & \textcolor{blue}{\textbf{0.320}} 0.340 & 0.352 0.346 \\
 Avg & \textcolor{red}{\textbf{0.225}} \textcolor{red}{\textbf{0.258}} & \textcolor{blue}{\textbf{0.226}} \textcolor{blue}{\textbf{0.262}} & 0.241 0.280 & 0.248 0.278 & 0.266 0.283 & 0.255 0.278 & 0.245 0.272 & 0.256 0.279 \\
 \midrule
 
\textbf{Exchange} & & & & & & & & \\
 96 & \textcolor{red}{\textbf{0.073(0.00)}} \textcolor{red}{\textbf{0.192(0.00)}} & \textcolor{blue}{\textbf{0.085}} 0.214 & 0.086 0.206 & 0.087 0.206 & -- -- & -- -- & 0.091 0.211 & 0.088 \textcolor{blue}{\textbf{0.205}} \\
 192 & \textcolor{red}{\textbf{0.165(0.01)}} \textcolor{red}{\textbf{0.285(0.01)}} & \textcolor{blue}{\textbf{\underline{0.175}}} 0.313 & 0.182 0.305 & 0.181 0.302 & -- -- & -- -- & 0.186 0.305 & 0.176 \textcolor{blue}{\textbf{0.299}} \\
 336 & 0.320(0.01) \textcolor{blue}{\textbf{0.402(0.01)}} & 0.316 0.420 & 0.331 0.417 & 0.333 0.418 & -- -- & -- -- & 0.380 0.449 & \textcolor{red}{\textbf{0.301}} \textcolor{red}{\textbf{0.397}} \\
 720 & \textcolor{red}{\textbf{0.830(0.02)}} \textcolor{red}{\textbf{0.671(0.02)}} & 0.851 \textcolor{blue}{\textbf{\underline{0.689}}} & \textcolor{blue}{\textbf{\underline{0.831}}} 0.691 & 0.830 0.685 & -- -- & -- -- & 0.896 0.712 & 0.901 0.714 \\
 Avg & \textcolor{red}{\textbf{0.347}} \textcolor{red}{\textbf{0.388}} & \textcolor{blue}{\textbf{0.357}} \textcolor{blue}{\textbf{0.391}} & \textcolor{blue}{\textbf{0.357}} 0.405 & 0.358 0.403 & -- -- & -- -- & 0.388 0.419 & 0.367 0.404 \\
 \midrule
 \textbf{1st Count} & \textbf{23 \ \ \ 24} & \textbf{12 \ \ \ 13} & \textbf{0 \ \ \ 0} & \textbf{0 \ \ \ 0} & \textbf{0 \ \ \ 0} & \textbf{4 \ \ \ 2} & \textbf{0 \ \ \ 0} & \textbf{1 \ \ \ 1} \\
 \midrule
\textbf{Average} & \textcolor{blue}{\textbf{0.328}} \ \textcolor{red}{\textbf{0.335}} & \textcolor{red}{\textbf{0.320}} \ 0.337 & 0.329 \ 0.344 & 0.338 \ 0.349 & 0.382 \ 0.366 & 0.334 \ \textcolor{blue}{\textbf{0.340}} & 0.345 \ 0.355 & 0.348 \ 0.354 \\
\midrule
\textbf{Impr. \%} & -- \ \ -- & -2.50 \ 0.59 & 0.30 \ 2.62 & 2.96 \ 4.01 & 14.14 \ 8.47 & 1.80 \ 1.47 & 4.93 \ 5.63 & 5.75 \ 5.37 \\
\bottomrule
\end{tabular}
\end{table*}

Long-term forecasting involves predicting future values over extended horizons, ranging from hours to days ahead. Following the standard protocol, we evaluated RhyMix on eight benchmark datasets with $L=96$ and $H \in \{96, 192, 336, 720\}$. Table~\ref{tab:forecasting_comparison} presents results against recent baselines.

RhyMix achieves improved performance across most datasets. On ETT datasets, it achieves an average MSE/MAE of 0.421/0.417 (ETTh1), 0.356/0.382 (ETTh2), 0.366/ 0.380 (ETTm1), and 0.262/0.310 (ETTm2), with lower errors than Time-o1, SOFTS, and TexFilter. On ECL, it achieves 0.160/0.251, with a 9.6\% reduction over SOFTS at H=720. On Weather, it achieves 0.225/0.258, with lower errors than all baselines. On Exchange, it achieves 0.347/0.388, with better performance at H=96 (0.073/0.192) across all baselines. On Traffic (862 sensors), RhyMix achieves 0.486/0.295, ranking fourth behind SOFTS, TimeMixer++, and Time-o1. Traffic data contains strong spatial inter-sensor dependencies that benefit CD-specialized architectures; we discuss this limitation explicitly in Sections~\ref{failure_analysis}.


To further validate on traffic tasks, we evaluate on four PEMS datasets (PEMS03, PEMS04, PEMS07, PEMS08) with $H \in \{12, 24, 48, 96\}$ (Table~\ref{tab:pems_results}). RhyMix achieves improved first/second-rank performance across all datasets. On PEMS03, it achieves 0.136/ 0.239, with lower errors than MSTN- Transformer by 12.8\% and TSMixer by 16.0\%. On PEMS04, it achieves 0.109/0.228, comparable to MSTN-Transformer (0.115 /0.226). On PEMS07, it achieves 0.127/0.233, with lower errors than MSTN-Transformer by 14.8\%. On PEMS08, it achieves 0.181/0.264, lower than all baselines.
    
\begin{table}
\centering
\scriptsize
\caption{Long-term forecasting evaluation results on PEMS datasets. Prediction horizons $H \in \{12, 24, 48, 96\}$. \textcolor{red}{Red}/\textcolor{blue}{Blue}: First/Second ranks. We evaluate each model five times and report the mean (standard deviation); results marked with \underline{underline} are not significantly worse than the first rank (Wilcoxon signed-rank test with Holm–Bonferroni correction, $\alpha = 0.05$). Example: 0.010(0.001) means 0.010 $\pm$ 0.001. Tra.: Transformer, BiL.: BiLSTM, iTrans.: iTransformer, M.TCN: ModernTCN, Impr.: Improvement \%. }
\label{tab:pems_results}
\setlength{\tabcolsep}{0.2pt}
\begin{tabular}{@{}lcccccccc@{}}
\toprule
\textbf{Dataset/H} & \textbf{RhyMix} & \textbf{MSTN-Tra.} & \textbf{MSTN-BiL.} & \textbf{iTrans.} & \textbf{M.TCN} & \textbf{PatchTST} & \textbf{TSMixer} & \textbf{Autoformer} \\
   
\textbf{H-Metric} & \textbf{MSE MAE} & \textbf{MSE MAE} & \textbf{MSE MAE} & \textbf{MSE MAE} & \textbf{MSE MAE} & \textbf{MSE MAE} & \textbf{MSE MAE} & \textbf{MSE MAE} \\
\midrule
    
\multicolumn{9}{l}{\textbf{PEMS03}} \\
12 & \textcolor{red}{\textbf{0.064(0.00)}} \textcolor{red}{\textbf{0.170(0.00)}} & 0.131 0.237 & 0.141 0.249 & \textcolor{blue}{\textbf{\underline{0.069}}} \textcolor{blue}{\textbf{\underline{0.175}}} & 0.112 0.221 & 0.079 0.187 & 0.075 0.187 & 0.277 0.387 \\
24 & \textcolor{red}{\textbf{0.091(0.01)}} \textcolor{red}{\textbf{0.204(0.01)}} & 0.135 0.241 & 0.151 0.259 & \textcolor{blue}{\textbf{\underline{0.098}}} \textcolor{blue}{\textbf{\underline{0.209}}} & 0.173 0.281 & 0.124 0.235 & 0.113 0.238 & 0.422 0.466 \\
48 & \textcolor{red}{\textbf{0.157(0.02)}} \textcolor{red}{\textbf{0.256(0.02)}} & \textcolor{blue}{\textbf{\underline{0.174}}} \textcolor{blue}{\textbf{\underline{0.268}}} & \underline{0.178} 0.280 & 0.448 0.416 & 0.307 0.395 & 0.223 0.319 & 0.195 0.320 & 0.806 0.679 \\
96 & \textcolor{blue}{\textbf{0.232(0.02)}} 0.326(0.02) & \textcolor{red}{\textbf{0.182}} \textcolor{red}{\textbf{0.281}} & 0.197 \textcolor{blue}{\textbf{0.299}} & 1.215 0.831 & 1.041 0.779 & 0.368 0.425 & 0.266 0.380 & 0.710 0.634 \\
\textbf{Avg} & \textcolor{red}{\textbf{0.136}} \textcolor{red}{\textbf{0.239}} & \textcolor{blue}{\textbf{0.156}} \textcolor{blue}{\textbf{0.257}} & 0.167 0.272 & 0.458 0.408 & 0.408 0.419 & 0.199 0.291 & 0.162 0.281 & 0.554 0.542 \\
\midrule
    
\multicolumn{9}{l}{\textbf{PEMS04}} \\
12 & \textcolor{red}{\textbf{0.062(0.00)}} \textcolor{red}{\textbf{0.171(0.01)}} & 0.105 0.210 & 0.114 0.222 & \textcolor{blue}{\textbf{0.081}} \textcolor{blue}{\textbf{0.188}} & 0.132 0.245 & 0.101 0.209 & 0.085 0.195 & 0.562 0.577 \\
24 & \textcolor{red}{\textbf{0.082(0.01)}} \textcolor{red}{\textbf{0.201(0.01)}} & \textcolor{blue}{\textbf{0.111}} \textcolor{blue}{\textbf{\underline{0.217}}} & 0.120 0.228 & 0.124 0.232 & 0.244 0.338 & 0.161 0.267 & 0.112 0.228 & 0.637 0.617 \\
48 & \textcolor{red}{\textbf{0.112(0.02)}} 0.232(0.02) & \textcolor{blue}{\textbf{\underline{0.119}}} \textcolor{red}{\textbf{0.229}} & 0.127 \textcolor{blue}{\textbf{\underline{0.239}}} & 0.135 0.248 & 0.452 0.482 & 0.294 0.369 & 0.159 0.278 & 1.002 0.775 \\
96 & 0.180(0.03) 0.307(0.03) & \textcolor{red}{\textbf{0.126}} \textcolor{red}{\textbf{0.235}} & \textcolor{blue}{\textbf{0.137}} \textcolor{blue}{\textbf{0.247}} & 0.169 0.280 & 1.127 0.818 & 0.507 0.505 & 0.190 0.313 & 0.853 0.708 \\
\textbf{Avg} & \textcolor{red}{\textbf{0.109}} \textcolor{blue}{\textbf{0.228}} & \textcolor{blue}{\textbf{0.115}} \textcolor{red}{\textbf{0.226}} & 0.125 0.240 & 0.127 0.237 & 0.488 0.471 & 0.266 0.338 & 0.136 0.254 & 0.764 0.669 \\
\midrule
    
\multicolumn{9}{l}{\textbf{PEMS07}} \\
12 & \textcolor{red}{\textbf{0.057(0.01)}} \textcolor{red}{\textbf{0.161(0.00)}} & 0.131 0.221 & 0.179 0.254 & \textcolor{blue}{\textbf{\underline{0.066}}} \textcolor{blue}{\textbf{\underline{0.164}}} & 0.085 0.196 & 0.076 0.180 & 0.070 0.177 & 0.201 0.330 \\
24 & \textcolor{red}{\textbf{0.080(0.01)}} \textcolor{red}{\textbf{0.184(0.01)}} & 0.142 0.232 & 0.183 0.260 & \textcolor{blue}{\textbf{\underline{0.087}}} \textcolor{blue}{\textbf{\underline{0.190}}} & 0.127 0.245 & 0.127 0.234 & 0.105 0.221 & 0.304 0.402 \\
48 & \textcolor{red}{\textbf{0.125(0.02)}} \textcolor{red}{\textbf{0.237(0.02)}} & \textcolor{blue}{\textbf{\underline{0.155}}} \textcolor{blue}{\textbf{\underline{0.241}}} & 0.189 0.265 & 0.892 0.764 & 0.267 0.380 & 0.238 0.325 & 0.157 0.265 & 0.422 0.472 \\
96 & \textcolor{blue}{\textbf{0.190(0.03)}} 0.285(0.03) & \textcolor{red}{\textbf{0.168}} \textcolor{red}{\textbf{0.252}} & 0.196 \textcolor{blue}{\textbf{0.273}} & 0.972 0.789 & 0.736 0.673 & 0.394 0.432 & 0.268 0.342 & 0.519 0.546 \\
\textbf{Avg} & \textcolor{red}{\textbf{0.127}} \textcolor{red}{\textbf{0.233}} & \textcolor{blue}{\textbf{0.149}} \textcolor{blue}{\textbf{0.237}} & 0.187 0.263 & 0.504 0.477 & 0.304 0.374 & 0.209 0.293 & 0.150 0.251 & 0.362 0.438 \\
\midrule
    
\multicolumn{9}{l}{\textbf{PEMS08}} \\
12 & \textcolor{red}{\textbf{0.075(0.00)}} \textcolor{red}{\textbf{0.178(0.01)}} & 0.291 0.297 & 0.309 0.314 & \textcolor{blue}{\textbf{\underline{0.089}}} \textcolor{blue}{\textbf{\underline{0.193}}} & 0.125 0.239 & 0.091 0.195 & 0.095 0.203 & 0.467 0.503 \\
24 & \textcolor{red}{\textbf{0.119(0.02)}} \textcolor{red}{\textbf{0.218(0.02)}} & 0.287 0.299 & 0.342 0.340 & \textcolor{blue}{\textbf{0.138}} \textcolor{blue}{\textbf{0.243}} & 0.238 0.336 & 0.144 0.247 & 0.150 0.257 & 0.503 0.512 \\
48 & \textcolor{red}{\textbf{0.201(0.02)}} \textcolor{blue}{\textbf{0.279(0.02)}} & 0.333 0.328 & 0.344 0.346 & \textcolor{blue}{\textbf{0.237}} \textcolor{red}{\textbf{0.277}} & 0.528 0.534 & 0.254 0.332 & 0.256 0.344 & 0.964 0.729 \\
96 & \textcolor{red}{\textbf{0.330(0.03)}} 0.380(0.03) & 0.362 \textcolor{blue}{\textbf{0.364}} & 0.398 0.393 & \textcolor{blue}{\textbf{\underline{0.346}}} \textcolor{red}{\textbf{0.363}} & 1.150 0.808 & 0.435 0.441 & 0.399 0.415 & 1.021 0.763 \\
\textbf{Avg} & \textcolor{red}{\textbf{0.181}} \textcolor{red}{\textbf{0.264}} & 0.318 0.322 & 0.348 0.348 & \textcolor{blue}{\textbf{0.202}} \textcolor{blue}{\textbf{0.269}} & 0.510 0.479 & 0.231 0.304 & 0.225 0.305 & 0.739 0.627 \\
\midrule
\textbf{1st Count} & \textbf{17 \ \ \ 13} & \textbf{3 \ \ \ 5} & \textbf{0 \ \ \ 0} & \textbf{0 \ \ \ 2} & \textbf{0 \ \ \ 0} & \textbf{0 \ \ \ 0} & \textbf{0 \ \ \ 0} & \textbf{0 \ \ \ 0} \\
\midrule
\textbf{Average} & \textcolor{red}{\textbf{0.138}} \ \textcolor{red}{\textbf{0.241}} & 0.185 \ \textcolor{blue}{\textbf{0.261}} & 0.207 \ 0.281 & 0.323 \ 0.348 & 0.428 \ 0.436 & 0.226 \ 0.307 & \textcolor{blue}{\textbf{0.168}} \ 0.273 & 0.605 \ 0.569 \\

\midrule
\textbf{Imp. \%} & -- \ -- & 25.41 \ 7.66 & 33.33 \ 14.23 & 57.28 \ 30.75 & 67.76 \ 44.72 & 38.94 \ 21.50 & 17.86 \ 11.72 & 77.19 \ 57.65 \\
\bottomrule
\end{tabular}
\end{table}
    
    \begin{figure}[t]
\centering
\includegraphics[width=0.95\textwidth]{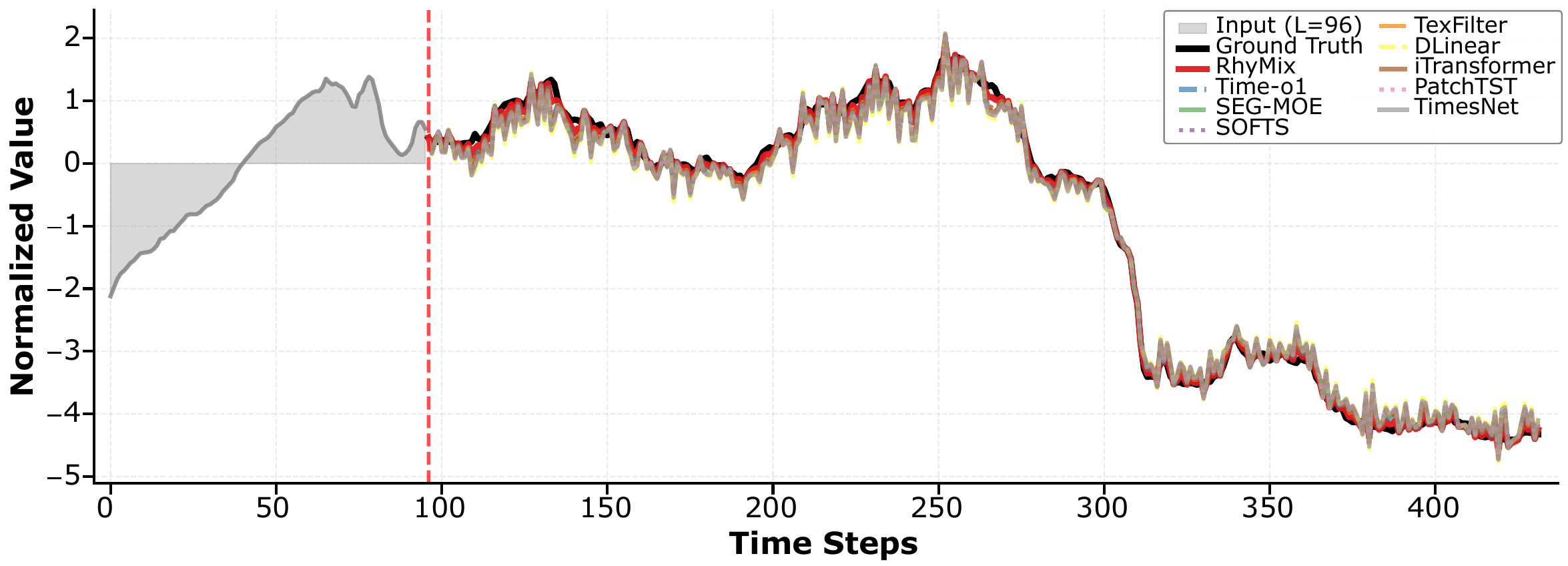}
\caption{Visualization of Weather dataset predictions by different models under the input-96-predict-336.}
\label{fig:weather}
\end{figure}
        
\begin{figure}
\centering
\includegraphics[width=0.95\textwidth]{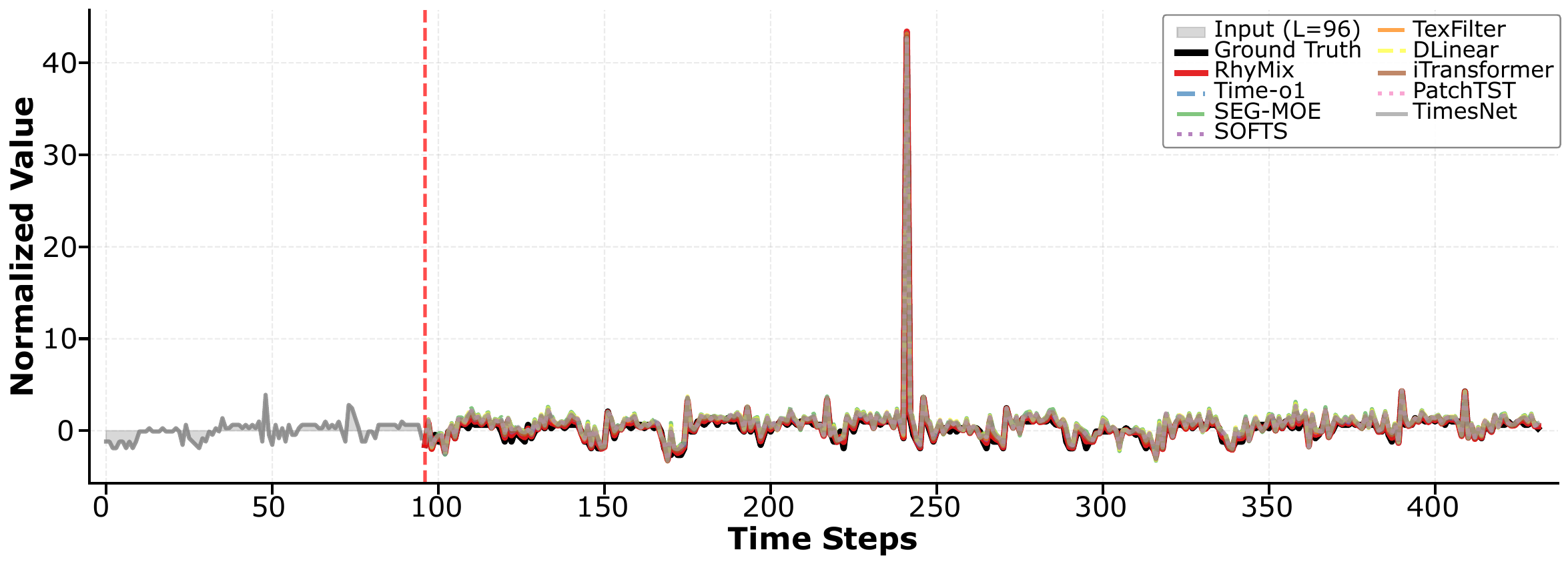}
\caption{Visualization of ECL dataset predictions by different models under the input-96-predict-336.}
\label{fig:ecl}
\end{figure}

Figures~\ref{fig:weather} and~\ref{fig:ecl} visualize forecasting on Weather and ECL datasets. On Weather (Fig.~\ref{fig:weather}), RhyMix achieves 0.243/0.272 at H=336, capturing daily cycles and seasonal trends with stable forecasts. On ECL (Fig.~\ref{fig:ecl}), RhyMix achieves 0.134/0.221 at H=96 and 0.160/0.251 at H=336, demonstrating robust performance on high-dimensional electricity data with 321 channels. RhyMix's predictions follow observed consumption patterns, with peaks and troughs at close temporal positions.

\subsection{Impact of Lookback Window Size}
\label{sec:lookback_analysis}

\begin{figure}
\centering
\includegraphics[width=0.97\textwidth]{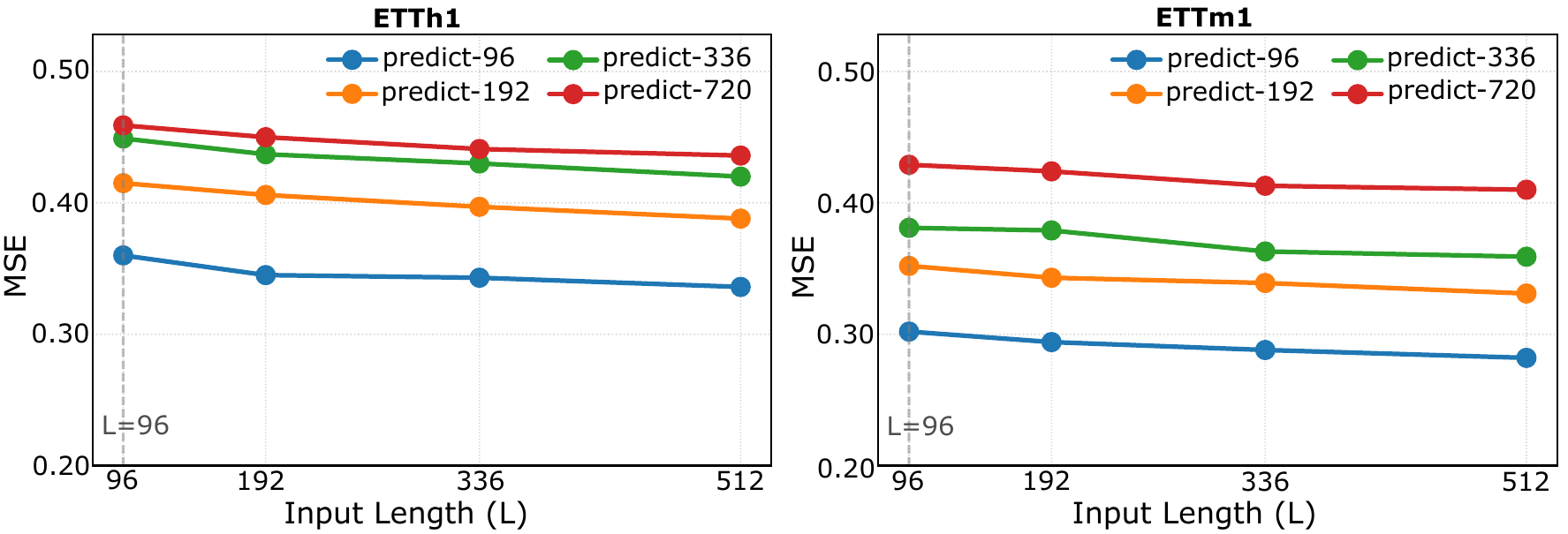}
\caption{Impact of input length $L$ on MSE across ETTh1 and ETTm1 dataset.}
\label{fig:lookback_sensitivity}
\end{figure}

To assess the sensitivity of RhyMix to the lookback window length, we evaluate the model with $L \in \{96, 192, 336, 512\}$ on the ETTh1 and ETTm1 datasets. Fig.~\ref{fig:lookback_sensitivity} visualizes the MSE across different input lengths for each dataset and prediction horizon $H \in \{96, 192, 336, 720\}$. Longer lookback windows generally reduce error.

\subsection{Visualization of RhyMix Signal Flow}

\subsubsection{Wave Representation of RhyMix Components}
\label{sec:wave_representation}

\begin{figure}[t]
\centering
\includegraphics[width=\textwidth]{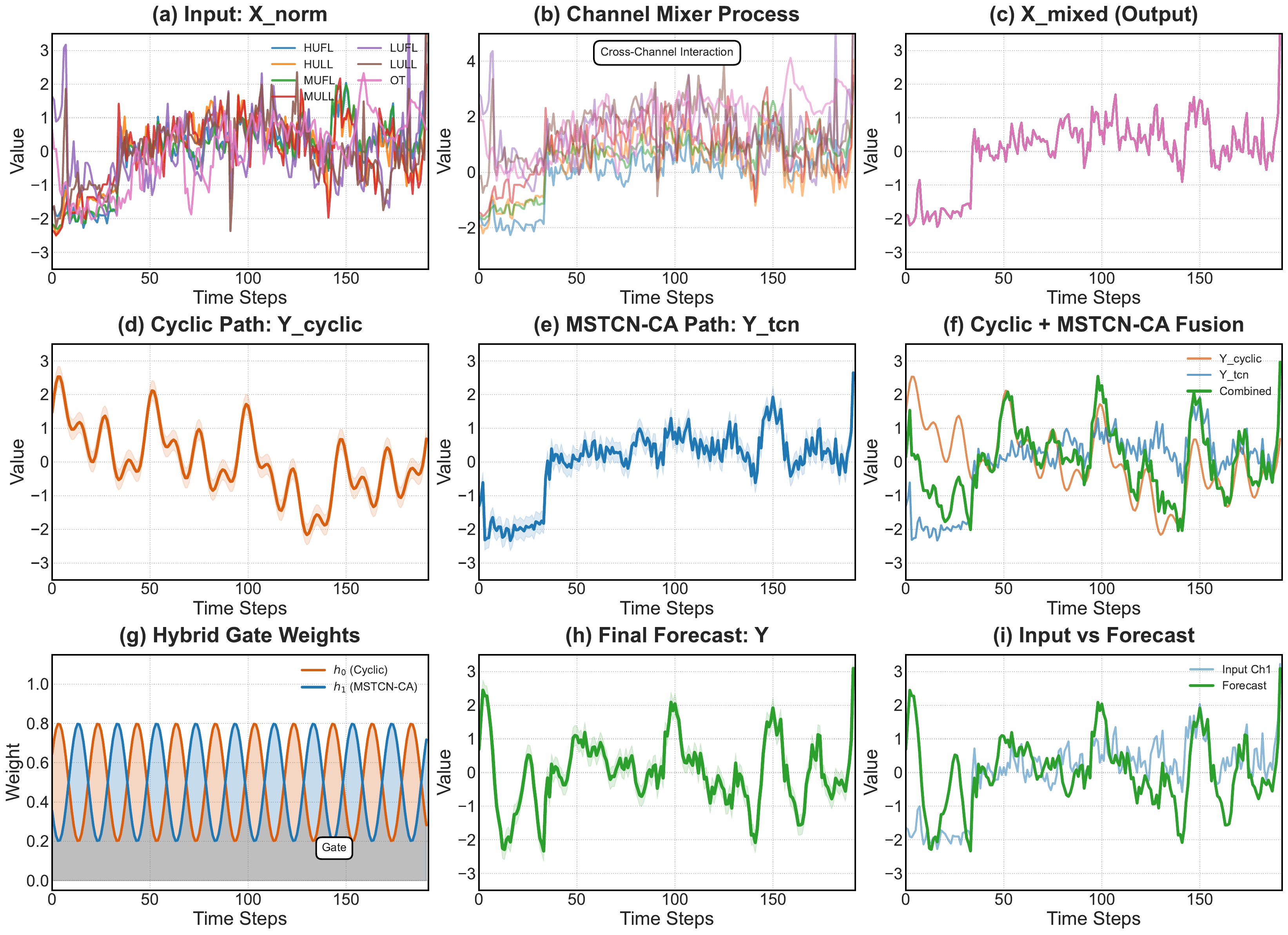}
\caption{Wave representations at key stages of RhyMix (On ETTh1 dataset, $C=7$)}
\label{fig:rhymix_waves}
\end{figure}

Fig.~\ref{fig:rhymix_waves} visualizes the information flow in RhyMix at key stages of the architecture using the ETTh1 dataset ($C=7$ channels). As shown in Fig.~\ref{fig:rhymix_waves}(a), the input channels exhibit diverse temporal characteristics, including different frequencies and amplitudes. The Channel Mixer (Fig.~\ref{fig:rhymix_waves}(b)) applies a bottleneck MLP ($C \rightarrow r \rightarrow C$ with $r = \min(64, \max(8, C//8))$) for cross-channel interaction (Section~\ref{sec:mathematical}, 
Eq.~2). For ETTh1 ($C=7$), this is $7 \rightarrow 8 \rightarrow 7$, producing $X_{\text{mixed}}$ (Fig.~\ref{fig:rhymix_waves}(c)). This mixed representation is then processed through two complementary paths. The Cyclic Path (Fig.~\ref{fig:rhymix_waves}(d)) extracts smooth seasonal patterns using explicit periodic embeddings with periods $P \in \{12, 24, 48, 168\}$, while the MSTCN-CA Path (Fig.~\ref{fig:rhymix_waves}(e)) captures finer multi-scale dynamics through dilated convolutions. The outputs of both paths are combined (Fig.~\ref{fig:rhymix_waves}(f)) and fused adaptively via the hybrid gate (Fig.~\ref{fig:rhymix_waves}(g)), which learns sample-specific weights $h_0$ and $h_1$ to balance the contributions of seasonal and multi-scale features (Sections~\ref{Arch_overview} and~\ref{sec:mathematical}). The final forecast (Fig.~\ref{fig:rhymix_waves}(h)) demonstrates the model's ability to capture both smooth trends and fine-grained variations, as confirmed by the input-forecast comparison (Fig.~\ref{fig:rhymix_waves}(i)). This visualization shows RhyMix captures both seasonal and multi-scale patterns, with the hybrid gate adaptively weighting each path's contributions to produce accurate forecasts across diverse temporal dynamics. The learned period weights $\lambda_k$ enable the model to emphasize different periodicities depending on the input, complementing the qualitative visualization  in Fig.~\ref{fig:rhymix_waves}. The quantitative contribution of the periodic 
features is confirmed by the ablation study in Section~\ref{sec:Ablation}, where 
removing them causes the largest MSE degradation on the four main datasets 
(ETTh1, ETTh2, Weather, ECL).

\subsubsection{Channel Mixer Wave Visualization}
\label{sec:channel_mixer_wave}

\begin{figure}[t]
\centering
\includegraphics[width=\textwidth]{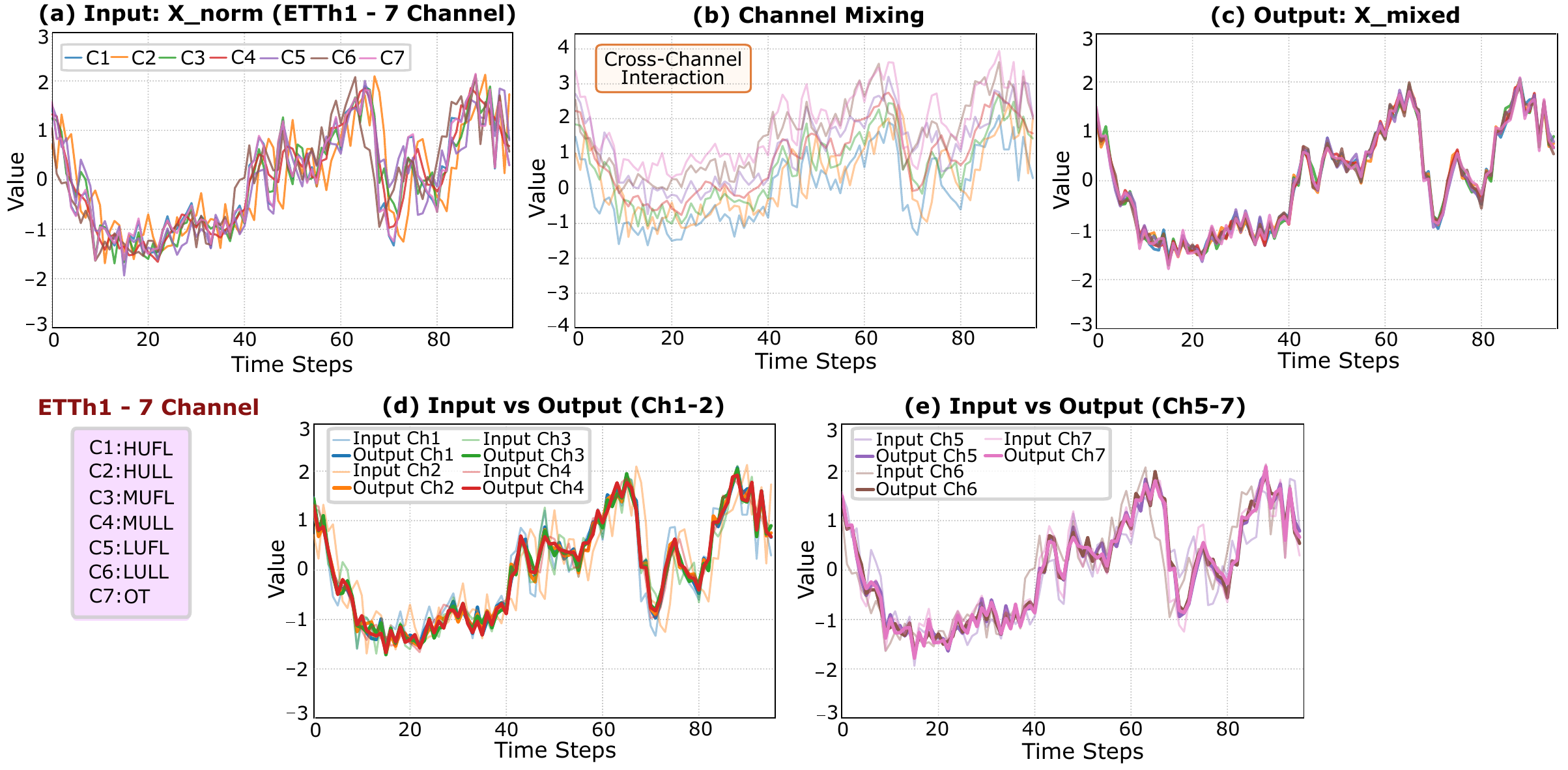}
\caption{Channel mixer: input-to-output wave visualization (On ETTh1 dataset, $C=7$).}
\label{fig:channel_mixer}
\end{figure}

Fig.~\ref{fig:channel_mixer} visualizes the transformation of input channels through the channel mixer process. As shown in Fig.~\ref{fig:channel_mixer}(a), the input channels exhibit diverse temporal characteristics, including different frequencies and amplitudes. During the mixing process (Fig.~\ref{fig:channel_mixer}(b)), channels interact through the bottleneck MLP, enabling information flow across variates. The resulting mixed representation $X_{\text{mixed}}$ (Fig.~\ref{fig:channel_mixer}(c)) preserves the number of channels while incorporating cross-channel dependencies. The input-output comparisons (Fig.~\ref{fig:channel_mixer}(d-e)) reveal how each channel is transformed by incorporating information from other channels, confirming effective cross-channel interaction.

\subsection{Ablation Study}
\label{sec:Ablation}

\begin{table*}[t]
\centering
\scriptsize
\caption{Ablation study on ETTh1, ETTh2, Weather, and ECL datasets ($H \in \{96, 192, 336, 720\}$). \textcolor{red}{\textbf{Red}}/\textcolor{blue}{\textbf{Blue}}: First/Second ranks. w/o A. Gates: Adaptive Gates}
\label{tab:ablation_Long_term}
\setlength{\tabcolsep}{2pt}
\begin{tabular}{@{}l|c|c|c|c|c|c|c@{}}
\toprule
\textbf{Dataset} & \textbf{Horizon} & \textbf{Full Model} & \textbf{w/o MSTCN-CA} & \textbf{w/o Cyclic Path} & \textbf{w/o A. Gates} & \textbf{w/o Periodic} & \textbf{w/o Local Conv} \\
& & \textbf{MSE MAE} & \textbf{MSE MAE} & \textbf{MSE MAE} & \textbf{MSE MAE} & \textbf{MSE MAE} & \textbf{MSE MAE} \\
\midrule
\textbf{ETTh1} & 96 & \textcolor{red}{\textbf{0.360}} \textcolor{red}{\textbf{0.382}} & 0.380 0.390 & \textcolor{blue}{\textbf{0.375}} 0.391 & 0.381 0.391 & 0.382 0.392 & 0.379 \textcolor{blue}{\textbf{0.388}} \\
& 192 & \textcolor{red}{\textbf{0.415}} \textcolor{red}{\textbf{0.411}} & 0.435 0.424 & \textcolor{blue}{\textbf{0.427}} \textcolor{blue}{\textbf{0.420}} & 0.435 0.422 & 0.443 0.427 & 0.433 0.423 \\
& 336 & \textcolor{red}{\textbf{0.449}} \textcolor{red}{\textbf{0.429}} & 0.470 \textcolor{blue}{\textbf{0.439}} & \textcolor{blue}{\textbf{0.469}} \textcolor{blue}{\textbf{0.439}} & 0.474 0.442 & 0.521 0.478 & 0.471 \textcolor{blue}{\textbf{0.439}} \\
& 720 & \textcolor{red}{\textbf{0.459}} \textcolor{red}{\textbf{0.445}} & 0.488 0.463 & 0.483 0.461 & 0.481 0.461 & 0.482 0.463 & \textcolor{blue}{\textbf{0.471}} \textcolor{blue}{\textbf{0.456}} \\
\midrule
\textbf{ETTh2} & 96 & \textcolor{red}{\textbf{0.269}} \textcolor{red}{\textbf{0.319}} & 0.290 0.338 & \textcolor{blue}{\textbf{0.289}} \textcolor{blue}{\textbf{0.337}} & 0.291 0.338 & 0.292 0.341 & \textcolor{blue}{\textbf{0.289}} \textcolor{blue}{\textbf{0.337}} \\
& 192 & \textcolor{red}{\textbf{0.349}} \textcolor{red}{\textbf{0.369}} & \textcolor{blue}{\textbf{0.371}} \textcolor{blue}{\textbf{0.390}} & 0.374 0.395 & 0.383 0.395 & 0.380 0.393 & 0.376 0.392 \\
& 336 & \textcolor{red}{\textbf{0.397}} \textcolor{red}{\textbf{0.411}} & 0.418 0.431 & 0.429 0.439 & 0.421 0.433 & \textcolor{blue}{\textbf{0.414}} 0.431 & 0.419 \textcolor{blue}{\textbf{0.427}} \\
& 720 & \textcolor{red}{\textbf{0.410}} \textcolor{red}{\textbf{0.429}} & 0.450 0.458 & 0.461 0.457 & 0.472 0.465 & 0.464 0.462 & \textcolor{blue}{\textbf{0.438}} \textcolor{blue}{\textbf{0.451}} \\
\midrule

\textbf{Weather} & 96 & \textcolor{red}{\textbf{0.142}} \textcolor{red}{\textbf{0.189}} & 0.159 0.206 & \textcolor{blue}{\textbf{0.158}} \textcolor{blue}{\textbf{0.205}} & \textcolor{blue}{\textbf{0.158}} 0.206 & 0.164 0.212 & 0.159 0.206 \\
& 192 & \textcolor{red}{\textbf{0.194}} \textcolor{red}{\textbf{0.241}} & \textcolor{blue}{\textbf{0.206}} 0.248 & 0.209 0.250 & 0.208 0.251 & 0.209 0.250 & 0.207 \textcolor{blue}{\textbf{0.246}} \\
& 336 & \textcolor{red}{\textbf{0.243}} \textcolor{red}{\textbf{0.272}} & 0.267 \textcolor{blue}{\textbf{0.290}} & 0.268 0.292 & 0.268 0.292 & 0.268 0.292 & \textcolor{blue}{\textbf{0.265}} \textcolor{blue}{\textbf{0.290}} \\
& 720 & \textcolor{red}{\textbf{0.321}} \textcolor{red}{\textbf{0.329}} & \textcolor{blue}{\textbf{0.349}} 0.346 & \textcolor{blue}{\textbf{0.349}} \textcolor{blue}{\textbf{0.345}} & 0.354 0.347 & 0.352 0.346 & 0.350 0.347 \\
\midrule
\textbf{ECL} & 96 & \textcolor{red}{\textbf{0.134}} \textcolor{red}{\textbf{0.221}} & 0.147 0.249 & \textcolor{blue}{\textbf{0.145}} 0.245 & 0.151 0.254 & 0.173 0.269 & 0.146 \textcolor{blue}{\textbf{0.244}} \\
& 192 & \textcolor{red}{\textbf{0.145}} \textcolor{red}{\textbf{0.243}} & 0.164 0.265 & 0.160 0.260 & 0.163 0.262 & 0.186 0.281 & \textcolor{blue}{\textbf{0.159}} \textcolor{blue}{\textbf{0.256}} \\
& 336 & \textcolor{red}{\textbf{0.160}} \textcolor{red}{\textbf{0.251}} & \textcolor{blue}{\textbf{0.175}} \textcolor{blue}{\textbf{0.277}} & 0.176 \textcolor{blue}{\textbf{0.277}} & 0.177 0.279 & 0.200 0.295 & 0.186 0.287 \\
& 720 & \textcolor{red}{\textbf{0.197}} \textcolor{red}{\textbf{0.287}} & \textcolor{blue}{\textbf{0.204}} \textcolor{blue}{\textbf{0.298}} & 0.210 0.305 & 0.211 0.306 & 0.228 0.316 & 0.207 0.299 \\
\bottomrule
\end{tabular}
\end{table*}

To evaluate each architectural component, we systematically remove individual modules from RhyMix, examining five variants: (i) \textit{w/o MSTCN-CA} removes the Contextual Forecasting Path; (ii) \textit{w/o Cyclic Path} removes the Cyclic Forecasting Path; (iii) \textit{w/o Adaptive Gates} replaces learned fusion weights with equal weights; (iv) \textit{w/o Periodic} removes all periodic features; and (v) \textit{w/o Local Conv} removes local convolutional.

The full model consistently outperforms all variants, confirming synergistic component interaction (Tables~\ref{tab:ablation_Long_term} and \ref{tab:ablation_pems}). Periodic features are most critical on the four main datasets, with removal causing the largest degradation across ETTh1, ETTh2, Weather, and ECL. On PEMS datasets, the local convolution branch is most critical. For example, on ECL at H=96, removing periodic features increases MSE from 0.134 to 0.173, while on ETTh1 at H=336, MSE rises from 0.449 to 0.521. The MSTCN-CA and Cyclic paths demonstrate complementary strengths: the Cyclic path suffices for ECL with periodic patterns (w/o MSTCN-CA achieves 0.204 vs. 0.197 for the full model at H=720), while MSTCN-CA is critical for ETTh1 with complex patterns (w/o MSTCN-CA increases MSE from 0.459 to 0.488 at H=720). Removing adaptive gates yields consistent degradation, confirming learned fusion weights are superior to fixed equal weighting.

\begin{table*}
\centering
\scriptsize
\caption{Ablation study on PEMS datasets across prediction horizons $H \in \{12, 24, 48, 96\}$. \textcolor{red}{\textbf{Red}} indicates first rank performance, \textcolor{blue}{\textbf{Blue}} indicates the second rank. w/o A. Gates: Adaptive Gates}
\label{tab:ablation_pems}
\setlength{\tabcolsep}{3pt}
\begin{tabular}{@{}l|c|c|c|c|c|c|c@{}}
\toprule
\textbf{Dataset} & \textbf{Horizon} & \textbf{Full Model} & \textbf{w/o MSTCN-CA} & \textbf{w/o Cyclic Path} & \textbf{w/o A. Gates} & \textbf{w/o Periodic} & \textbf{w/o Local Conv} \\
& & \textbf{MSE MAE} & \textbf{MSE MAE} & \textbf{MSE MAE} & \textbf{MSE MAE} & \textbf{MSE MAE} & \textbf{MSE MAE} \\
\midrule

\textbf{PEMS03} & 12 & \textcolor{red}{\textbf{0.064}} \textcolor{red}{\textbf{0.170}} & 0.073 0.184 & 0.076 0.188 & 0.075 0.185 & 0.074 0.185 & \textcolor{blue}{\textbf{0.073}} \textcolor{blue}{\textbf{0.184}} \\
& 24 & \textcolor{red}{\textbf{0.091}} \textcolor{red}{\textbf{0.204}} & 0.108 0.224 & 0.106 0.222 & 0.105 0.221 & \textcolor{blue}{\textbf{0.103}} \textcolor{blue}{\textbf{0.206}} & 0.104 0.220 \\
& 48 & \textcolor{red}{\textbf{0.157}} \textcolor{red}{\textbf{0.256}} & 0.182 0.292 & \textcolor{blue}{\textbf{0.168}} \textcolor{blue}{\textbf{0.281}} & 0.174 0.285 & 0.174 0.282 & 0.185 0.299 \\
& 96 & \textcolor{red}{\textbf{0.232}} \textcolor{red}{\textbf{0.326}} & 0.254 0.357 & \textcolor{blue}{\textbf{0.235}} 0.342 & 0.237 \textcolor{blue}{\textbf{0.338}} & 0.237 0.341 & 0.288 0.382 \\
\midrule

\textbf{PEMS04} & 12 & \textcolor{red}{\textbf{0.062}} \textcolor{red}{\textbf{0.171}} & 0.079 0.189 & 0.081 0.193 & \textcolor{blue}{\textbf{0.077}} \textcolor{blue}{\textbf{0.187}} & \textcolor{blue}{\textbf{0.077}} 0.190 & 0.085 0.197 \\
    & 24 & \textcolor{red}{\textbf{0.082}} \textcolor{red}{\textbf{0.201}} & 0.100 0.217 & 0.098 \textcolor{blue}{\textbf{0.214}} & \textcolor{blue}{\textbf{0.097}} 0.215 & 0.100 0.218 & 0.108 0.230 \\
    & 48 & \textcolor{red}{\textbf{0.112}} \textcolor{red}{\textbf{0.232}} & 0.147 0.273 & 0.143 0.269 & \textcolor{blue}{\textbf{0.141}} \textcolor{blue}{\textbf{0.265}} & 0.145 0.270 & 0.151 0.280 \\
    & 96 & \textcolor{red}{\textbf{0.180}} \textcolor{red}{\textbf{0.307}} & 0.207 0.335 & \textcolor{blue}{\textbf{0.197}} \textcolor{blue}{\textbf{0.326}} & 0.201 0.329 & 0.208 0.335 & 0.224 0.353 \\
\midrule

\textbf{PEMS07} & 12 & \textcolor{red}{\textbf{0.057}} \textcolor{red}{\textbf{0.161}} & 0.065 0.169 & 0.065 0.168 & \textcolor{blue}{\textbf{0.064}} 0.166 & \textcolor{red}{\textbf{0.057}} \textcolor{blue}{\textbf{0.162}} & 0.065 0.169 \\
& 24 & \textcolor{red}{\textbf{0.080}} \textcolor{red}{\textbf{0.184}} & \textcolor{blue}{\textbf{0.088}} \textcolor{blue}{\textbf{0.195}} & 0.089 0.197 & 0.091 0.199 & 0.091 0.199 & 0.092 0.205 \\
& 48 & \textcolor{red}{\textbf{0.125}} \textcolor{red}{\textbf{0.237}} & 0.140 0.253 & 0.139 0.254 & 0.144 0.259 & \textcolor{blue}{\textbf{0.135}} \textcolor{blue}{\textbf{0.248}} & 0.148 0.263 \\
& 96 & \textcolor{red}{\textbf{0.190}} \textcolor{red}{\textbf{0.285}} & 0.221 0.329 & \textcolor{blue}{\textbf{0.206}} \textcolor{blue}{\textbf{0.311}} & 0.217 0.322 & \textcolor{blue}{\textbf{0.206}} 0.313 & 0.259 0.361 \\
\midrule

\textbf{PEMS08} & 12 & \textcolor{red}{\textbf{0.075}} \textcolor{red}{\textbf{0.178}} & 0.089 0.197 & \textcolor{blue}{\textbf{0.086}} \textcolor{blue}{\textbf{0.195}} & 0.088 0.197 & 0.089 0.198 & 0.089 0.200 \\
& 24 & \textcolor{red}{\textbf{0.119}} \textcolor{red}{\textbf{0.218}} & 0.130 \textcolor{blue}{\textbf{0.241}} & \textcolor{blue}{\textbf{0.129}} \textcolor{blue}{\textbf{0.241}} & \textcolor{blue}{\textbf{0.129}} 0.242 & 0.130 0.243 & 0.131 0.246 \\
& 48 & \textcolor{red}{\textbf{0.201}} \textcolor{red}{\textbf{0.279}} & 0.217 0.321 & 0.219 0.317 & 0.219 \textcolor{blue}{\textbf{0.316}} & 0.224 0.319 & \textcolor{blue}{\textbf{0.215}} \textcolor{blue}{\textbf{0.317}} \\
& 96 & \textcolor{red}{\textbf{0.330}} \textcolor{red}{\textbf{0.380}} & 0.352 0.401 & \textcolor{blue}{\textbf{0.350}} \textcolor{blue}{\textbf{0.396}} & 0.408 0.430 & 0.373 0.415 & 0.367 0.407 \\
\bottomrule

\end{tabular}
\end{table*}

On PEMS traffic datasets, the local convolution branch is particularly critical. At H=96, removing local convolution increases MSE from 0.232 to 0.288 on PEMS03, from 0.180 to 0.224 on PEMS04, and from 0.190 to 0.259 on PEMS07. Periodic features also contribute meaningfully: on PEMS04 at H=12, removal increases MSE from 0.062 to 0.077, and on PEMS08 at H=96, MSE rises from 0.330 to 0.373. MSTCN-CA and Cyclic paths show complementary behavior, both contributing to performance.

\textbf{Progressive Bottom-Up Validation:} To verify every component is necessary, we evaluate a progressive construction of RhyMix on ECL (H=96). Starting from the cyclic-only baseline (w/o MSTCN-CA, MSE 0.147), adding MSTCN-CA improves MSE to 0.134. Removing local convolution (0.146), adaptive gates (0.151), or periodic features (0.173) all degrade performance, confirming each component contributes to the full model. On Weather (H=96), cyclic-only (0.159) improves with MSTCN-CA (0.142), while removing local convolution (0.159), adaptive gates (0.158), or periodic features (0.164) consistently degrades performance relative to the full model. On PEMS03 (H=96), the cyclic-only baseline (0.254) improves with MSTCN-CA (0.232), while removing adaptive gates (0.237), periodic features (0.237), or local convolution (0.288) degrades performance, confirming that all components contribute synergistically. In summary, the ablation study validates that each component contributes meaningfully. Periodic features emerge as most critical, while MSTCN-CA and cyclic paths provide complementary dataset-dependent capabilities. The full integrated model consistently delivers the best forecasting accuracy.

\begin{table}[t]
\centering
\scriptsize
\caption{Sensitivity of RhyMix to hyperparameters on ETTh1 ($L{=}H{=}96$). 
\textbf{Bold} marks the default configuration used throughout the main paper.}
\label{tab:sensitivity}
\setlength{\tabcolsep}{6pt}
\begin{tabular}{l|c|c}
\toprule
\textbf{Parameter} & \textbf{Value} & \textbf{ETTh1 (MSE)} \\
\midrule
\multirow{4}{*}{$P$ (period base)}
& 12 & 0.362 \\
& \textbf{24} & \textbf{0.360} \\
& 48 & 0.361 \\
& 96 & 0.365 \\
\midrule
\multirow{4}{*}{Dilations}
& (1,2,4) & 0.361 \\
& \textbf{(1,2,4,8)} & \textbf{0.360} \\
& (1,4,8,16) & 0.361 \\
& (1,2,4,8,16) & 0.362 \\
\midrule
\multirow{4}{*}{$r$ (bottleneck)}
& \textbf{8} & \textbf{0.360} \\
& 16 & 0.362 \\
& 32 & 0.364 \\
& 64 & 0.361 \\
\midrule
\multirow{3}{*}{$\delta$ (Huber)}
& 0.5 & 0.363 \\
& \textbf{1.0} & \textbf{0.360} \\
& 2.0 & 0.365 \\
\midrule
\multirow{4}{*}{Dropout}
& 0.0 & 0.361 \\
& \textbf{0.05} & \textbf{0.360} \\
& 0.1 & 0.367 \\
& 0.2 & 0.363 \\
\midrule
\multirow{4}{*}{LR}
& $1{\times}10^{-4}$ & 0.361 \\
& $3{\times}10^{-4}$ & 0.360 \\
& $\mathbf{7{\times}10^{-4}}$ & \textbf{0.360} \\
& $1{\times}10^{-3}$ & 0.362 \\
\bottomrule
\end{tabular}
\end{table}

Table~\ref{tab:sensitivity} reports sensitivity to six hyperparameters on ETTh1 ($L{=}H{=}96$). MSE varies within a narrow band (0.360--0.367) across 
the tested settings, indicating that RhyMix is moderately robust to hyperparameter choices. The default configuration used throughout the main paper achieves the lowest or tied-lowest MSE in every category, confirming that the chosen defaults are optimal or near-optimal.

\subsection{Model Complexity and Edge-AI Deployability}
\label{sec:model_complexity}

\subsubsection{Complexity Analysis}
\label{sec:complexity_analysis}

In this analysis, we denote the length of the look-back window as $L$, the number of variate features (channels) as $C$, and the prediction horizon as $H$. We simplify the asymptotic complexity expressions by omitting the model dimension $d$, assuming $d \approx C$.

RhyMix maintains linear complexity through its parallel dual-path design. The Cyclic Path uses explicit periodic embeddings with complexity $\mathcal{O}(L \cdot C \cdot K)$ ($K=4$ periods), while the MSTCN-CA Path employs depthwise dilated convolutions with complexity $\mathcal{O}(L \cdot C \cdot D)$ ($D=15$, sum of dilation rates 1+2+4+8). Both paths avoid the quadratic cost of self-attention. The total asymptotic complexity is dominated by the four forecasting heads, each applying a linear transformation from $L$ to $H$, resulting in $\mathcal{O}(L \cdot C \cdot (K + D + r + 4H))$, where $K=4$ is the number of cyclic periods, $D=15$ is the sum of dilation rates (1+2+4+8), and $r = \min(64, \max(8, \lfloor C/8 \rfloor))$ is the bottleneck dimension of the channel mixer. This scales linearly with both $L$ and $C$. RhyMix maintains a compact core of $\sim$40K parameters for standard configuration ($C=7$, $L=96$, $H=96$).

\textbf{Comparison with Time-o1 and TimeMixer++:} Time-o1~\citep{wang2026timeo} 
adds a projection matrix $P^* \in \mathbb{R}^{H \times H}$ requiring SVD pre-computation with $\mathcal{O}(m \cdot H^2)$ complexity and a base model. TimeMixer++~\citep{wang2025timemixergeneraltimeseries} introduces a multi-resolution pattern machine with complexity 
$\mathcal{O}(M \cdot C \cdot K \cdot (L \cdot f + p \cdot f))$ and parameter counts ranging from 647K to 4M. RhyMix offers two key advantages over Time-o1 
and TimeMixer++: (1) linear scaling with horizon ($\mathcal{O}(H)$) versus Time-o1's quadratic growth ($\mathcal{O}(H^2)$) and TimeMixer++'s multi-resolution imaging operations; and (2) no pre-computation or base model requirement. For low-dimensional datasets ($C \leq 21$), RhyMix also uses 
7–16$\times$ fewer parameters than Time-o1 and TimeMixer++. For high-dimensional datasets ($C > 300$), RhyMix's parameter count (314K–2.99M) is comparable to TimeMixer++ (647K–4M), but RhyMix retains linear complexity and requires no pre-computation or external base model.

\subsubsection{Comparative Analysis and Edge-AI Deployability}
\label{sec:comparative_edge_ai}

Table~\ref{tab:multirhythm_complexity_sota} compares RhyMix with recent SOTA models, ordered by parameter count on ETTh1 dataset. At $C=7$, RhyMix achieves the lowest parameter count (40,269) with linear complexity $\mathcal{O}(L \cdot C \cdot (K + D + r + 4H))$ and linear horizon scaling $\mathcal{O}(H)$. The compared models have parameter counts ranging from 50K (DLinear) to 4M (TimeMixer++), with complexity ranging from $\mathcal{O}(CL)$ to $\mathcal{O}(C^2 + CL + CH)$ or $\mathcal{O}(m \cdot H^2)$. At $C=7$, RhyMix uses 1.6$\times$ fewer parameters than CycleNet, 2.5$\times$ fewer than ModernTCN, and up to 16$\times$ fewer than TimeMixer++. Parameter count scales with $C$ and sampling frequency (Table~\ref{tab:param_counts}).

\begin{table}[t]
\centering
\scriptsize
\caption{Complexity and performance comparison between RhyMix and recent SOTA models (parameter count and ratio are shown at $C=7$ (standard 
configuration)). $\uparrow$ indicates the competitor model has more parameters than RhyMix.}
\label{tab:multirhythm_complexity_sota}
\begin{tabular}{l|l|c|c|c}
\toprule
\textbf{Model} & \textbf{Complexity} & \textbf{Horizon Scaling} & \textbf{Parameters} & \textbf{Ratio} \\
\midrule
DLinear  & $\mathcal{O}(CL)$ & Independent & $\sim$50K & \textbf{1.2$\times$ $\uparrow$} \\
CycleNet & $\mathcal{O}(L \cdot C)$ & $\mathcal{O}(H)$ & $\sim$65K & \textbf{1.6$\times$ $\uparrow$} \\
ModernTCN & $\mathcal{O}(L \cdot C)$ & $\mathcal{O}(H)$ & $\sim$100K & \textbf{2.5$\times$ $\uparrow$} \\
PatchTST (2023) & $\mathcal{O}(CL^2 + CH)$ & $\mathcal{O}(H)$ & $\sim$300K & \textbf{7.4$\times$ $\uparrow$} \\
SOFTS  & $\mathcal{O}(CL + CH)$ & $\mathcal{O}(H)$ & $\sim$300K & \textbf{7.4$\times$ $\uparrow$} \\
TimesNet  & $\mathcal{O}(CL \log L + CH)$ & $\mathcal{O}(H)$ & $\sim$350K & \textbf{8.7$\times$ $\uparrow$} \\
FilterNet / TexFilter  & $\mathcal{O}(CL \log L + CH)$ & $\mathcal{O}(H)$ & $\sim$350K & \textbf{8.7$\times$ $\uparrow$} \\
Time-o1  & $\mathcal{O}(m \cdot H^2)$ & $\mathcal{O}(H^2)$ & $\sim$290K & \textbf{7.2$\times$ $\uparrow$} \\
iTransformer & $\mathcal{O}(C^2 + CL + CH)$ & $\mathcal{O}(H)$ & $\sim$400K & \textbf{9.9$\times$ $\uparrow$} \\
SEG-MOE  & $\mathcal{O}(CL + CH + C^2)$ & $\mathcal{O}(H)$ & $\sim$500K & \textbf{12.4$\times$ $\uparrow$} \\
TimeMixer++  & $\mathcal{O}(M\cdot C\cdot K\cdot(L\cdot f + p\cdot f))$ & $\mathcal{O}(H)$ & $\sim$647K & \textbf{16$\times$ $\uparrow$} \\
\textbf{RhyMix (Ours)} & $\mathcal{O}(L \cdot C \cdot (K + D + r + 4H))$ & $\mathcal{O}(H)$ & \textbf{$\sim$40,269} & \textbf{1.0$\times$} \\
\bottomrule
\end{tabular}
\end{table}

\begin{table}[t]
\centering
\caption{Parameter count, model size, inference latency, and inference memory 
footprint of RhyMix across representative datasets 
($L=96$, $H=96$, batch size 32 for latency and memory). Values measured on 
NVIDIA A100.}
\label{tab:param_counts}
\scriptsize
\setlength{\tabcolsep}{4pt}
\begin{tabular}{@{}lccccc@{}}
\toprule
\textbf{Dataset} & \textbf{C} & \textbf{Parameters} & \textbf{Size (KB)} & \textbf{Latency (ms)} & \textbf{Inf. Mem. (MB)} \\
\midrule
ETTh1     & 7   & 40,269  & 157.3     & 5.85 & 10.9 \\
ETTm1     & 7   & 45,561  & 178.0     & 5.72 & 19.1 \\
Exchange  & 8   & 39,286  & 153.5     & 5.88 & 19.4 \\
Weather   & 21  & 71,965  & 281.1     & 6.02 & 24.5 \\
PEMS08    & 170 & 441,417 & 1{,}724.3 & 5.23 & 82.2 \\
PEMS04    & 307 & 840,079 & 3{,}281.6 & 7.24 & 135.2 \\
\bottomrule
\end{tabular}
\end{table}

Table~\ref{tab:param_counts} reports parameter count and resource usage for six representative datasets. The parameter count depends on both $C$ and the sampling frequency: higher-frequency datasets require larger cyclic periods, 
which increases the embedding table size even at lower $C$. This is why PEMS08 ($C=170$, 5-minute) has 441K parameters, more than Exchange ($C=8$, daily) at 39K, while within each sampling frequency parameter count scales linearly with $C$.

\textbf{Edge-AI Deployability:} To evaluate practical deployability, we measured 
inference time and memory footprint across representative datasets (Table~\ref{tab:param_counts}). RhyMix achieves an inference latency of 5.85 ms per batch (batch size 32) on ETTh1 ($C=7$) with 10.9 MB peak inference memory and a 157 KB model size. For high-dimensional datasets such as PEMS04 ($C=307$), latency rises to 7.24 ms with 135 MB peak inference memory and a 3.20 MB model size. In comparison, TimeMixer++ requires significantly more memory 
($>$500 MB) due to 2-D imaging operations, while Time-o1 requires additional memory for SVD pre-computation and the projection matrix $P^* \in \mathbb{R}^{H \times H}$. RhyMix requires no pre-computation, enabling instant deployment on edge devices. Inference time and memory scale with the 
number of channels, ensuring consistent performance across datasets. The compact footprint (157 KB for $C=7$) and sub-8 ms inference latency make RhyMix well-suited for resource-constrained and real-time Edge-AI applications.

    \subsection{Limitations and Failure Cases}
\label{failure_analysis}

RhyMix shows limitations on Traffic and PEMS04 datasets where specialized architectures achieve lower errors. On Traffic, RhyMix achieves 0.486/0.295, while SOFTS (0.409/0.267), iTransformer (0.428/0.282), and Time-o1 (0.419/0.280) achieve lower errors. Traffic data consists of 862 sensors with spatial dependencies, and RhyMix's hybrid CI+CD approach may not fully capture the intricate inter-sensor relationships that specialized CD models can model. However, RhyMix achieves lower errors than DLinear (0.625/0.383) by 22.2\%. On PEMS04, RhyMix achieves an average MSE/MAE of 0.109/0.228, with lower MSE than all baselines and competitive MAE (second best after MSTN-Transformer at 0.226). RhyMix performs better at horizons (H=12, 24, 48), but at H=96, MSTN-Transformer achieves lower MSE with 0.126/0.235 compared to RhyMix's 0.180/0.307. This suggests that RhyMix's adaptive gating mechanisms may be less effective for long-horizon predictions on certain datasets. A second limitation is the fixed phase offset $\phi=0$ used in the cyclic embedding lookup (Eq.~\ref{eq:cyclic}). The phase offset determines the starting position of each cycle relative to the input window. With $\phi=0$, the model always begins at position 0 of the embedding table, regardless of the actual phase alignment of the input window. Despite this fixed offset, RhyMix achieves state-of-the-art performance on ETTm1 and ETTm2, datasets with arbitrary window sampling offsets. This suggests that the learnable embedding tables compensate by encoding all cyclic phases across their $P_k$ entries rather than relying on phase alignment. Future work could explore learnable phase offsets, but the current design appears sufficient in practice.

These gaps reflect the trade-off in RhyMix's design: prioritizing general-purpose multi-scale learning with adaptive gating over specialized inductive biases. By focusing on broad applicability across heterogeneous scenarios, RhyMix achieves competitive performance on most benchmarks but may be outperformed on datasets where explicitly optimized architectures have advantages. This trade-off aligns with our design philosophy of a lightweight, efficient model for diverse real-world applications. The modular architecture allows practitioners to adapt it—for instance, by adjusting cyclic periods for different frequencies or incorporating domain-specific preprocessing—which are promising directions for future work.

  \section{Conclusions}

This work introduces RhyMix, a lightweight hybrid architecture combining explicit cyclic priors with data-driven multi-scale convolutions. RhyMix employs a dual-path design—a Cyclic Path for seasonal patterns and an MSTCN-CA Path for multi-scale dynamics—with adaptive gating that dynamically fuses four specialized forecasting heads per sample. Unlike methods relying on rigid assumptions, our approach integrates explicit cyclic priors with multi-scale TCNs, using adaptive gating to modulate feature contributions. This design captures diverse temporal patterns while maintaining linear complexity and delivering SOTA performance.

Across extensive evaluations, RhyMix achieves the best average MSE on 9 of 12 benchmark datasets and the best average MAE on 8 of 12 datasets. At $C=7$, RhyMix uses substantially fewer parameters (40K) than comparable methods such as Time-o1 ($\sim$290K) and TimeMixer++ ($\sim$647K); parameter count scales with channel count and sampling frequency. RhyMix achieves improved prediction performance across diverse domains, with an inference latency of 5.85 ms on ETTh1 ($C=7$) and a compact footprint of 157 KB. This simultaneous advancement in both performance and efficiency distinguishes RhyMix from prior work. Unlike methods requiring quadratic scaling, pre-computation, or external base models, RhyMix is fully self-contained with linear complexity, making it suitable for real-time edge deployment. Ablation studies confirm that each component contributes meaningfully, with periodic features providing foundational capability and the MSTCN-CA and cyclic paths delivering significant complementary gains.

Future work will explore extending RhyMix to other time series tasks such as imputation and anomaly detection, as well as investigating adaptive period selection mechanisms, such as learnable period lengths 
or FFT-based automatic period detection, to automatically discover optimal cyclic periods for different datasets. Our work paves the way for efficient multi-scale temporal modeling under resource constraints, enabling real-time applications across diverse domains from energy 
systems to intelligent transportation.
    
\section*{Data and Code Availability}
All datasets are publicly available from their respective sources as cited. The source code will be released upon acceptance.
    
\section*{CRediT Authorship Contribution Statement}
\textbf{Sumit Satishrao Shevtekar:} Conceptualization, Formal analysis, Investigation, Methodology, Software, Visualization, Validation, Writing – original draft. \textbf{Chandresh Kumar Maurya:} Conceptualization, Formal analysis, Investigation, Resources, Methodology, Validation, Supervision, Writing – review \& editing.
    
 \section*{Declaration of Competing Interest} 
    
The authors declare no conflict of interest.

     \bibliographystyle{elsarticle-num}

     \bibliography{cas-refs}
    
    
    
    \end{document}